\documentclass[11pt]{article}
\usepackage[english]{babel}
\usepackage[left=1in,right=1in,top=1in,bottom=1.5in]{geometry} 
\usepackage[skip=5pt, indent=10pt]{parskip}
\usepackage{amssymb,amsmath,amsthm}
\usepackage{xcolor}
\usepackage{nicefrac}
\usepackage[margin=1cm, font=small]{caption}
\usepackage{natbib}
\usepackage[makeroom]{cancel}
\usepackage{bbm}
\usepackage{mathrsfs}
\usepackage{comment}
\usepackage{algorithm}
\usepackage{algpseudocode}
\usepackage{times}
\usepackage{graphicx}
\usepackage{subfig}
\usepackage[most]{tcolorbox}
\usepackage{setspace}
\usepackage{makecell}
\usepackage{upgreek}
\usepackage{mathabx}
\usepackage{hyperref}

\definecolor{darkred}{HTML}{880000}
\definecolor{darkblue}{HTML}{000088}

\hypersetup{
  colorlinks,
  linkcolor={darkred},
  citecolor={darkblue}
}

\def\argmin{\operatornamewithlimits{argmin}}

\renewcommand{\leq}{\leqslant}
\renewcommand{\geq}{\geqslant}

\renewcommand{\ge}{\geqslant}

\newcommand{\ceil}[1]{\left\lceil #1 \right\rceil}

\newcommand{\dd}{{\rm d}}

\newcommand{\m}{\mathfrak m}

\newcommand{\R}{\mathbb {R}}
\newcommand{\N}{\mathbb {N}}
\newcommand{\E}{\mathbb{E}}
\newcommand{\p}{\mathbb{P}}

\newcommand{\ca}{{\mathcal{A}}}
\newcommand{\cb}{{\mathcal B}}

\newcommand{\cd}{{\mathcal D}}
\newcommand{\ce}{{\mathcal E}}
\newcommand{\cf}{{\mathcal F}}

\newcommand{\ch}{{\mathcal H}}

\newcommand{\cn}{{\mathcal N}}
\newcommand{\co}{\mathcal{O}}
\newcommand{\cp}{\mathcal{P}}

\newcommand{\rr}{\mathcal{R}}
\newcommand{\cs}{\mathcal{S}}
\newcommand{\ct}{\mathcal{T}}

\newcommand{\sfp}{\mathsf{p}}
\newcommand{\sfq}{\mathsf{q}}

\newcommand{\sfx}{\mathsf{X}}

\newcommand{\scp}{\mathscr{P}}
\newcommand{\scq}{\mathscr{Q}}
\newcommand{\scv}{\mathscr{V}}

\newcommand{\z}{\mathfrak{z}}

\newcommand{\tstar}{{\tau^*}}

\newcommand{\Var}{{\mathrm{Var}}}

\newcommand{\regret}{{\mathtt{Reg}}}
\newcommand{\mar}{{\mathtt{MAR}}}
\DeclareMathOperator{\JS}{\mathrm{JS}}
\DeclareMathOperator{\KL}{\mathrm{KL}}

\renewcommand{\leq}{\leqslant}
\renewcommand{\geq}{\geqslant}

\newcommand{\myendproof}{\hfill{\LARGE$\square$}}

\newtheorem{Th}{Theorem}[section]
\newtheorem{Alg}[Th]{Algorithm}

\newtheorem{Lem}[Th]{Lemma}

\newtheorem{Rem}[Th]{Remark}

\title{Noise-contrastive Online Change Point Detection\thanks{The preliminary version of this paper \citep{puchkin23} was presented at the 26th International Conference on Artificial Intelligence and Statistics (AISTATS), 2023.}}

\author{%
	Nikita Puchkin\thanks{HSE University, Russian Federation}
    \qquad\quad
	Artur Goldman\thanks{ETH Z\"urich, ETH AI Center, Switzerland}
    \qquad\quad
    Konstantin Yakovlev\footnotemark[2]
    \\[2ex]
	Valeriia Dzis\footnotemark[2]
    \qquad\quad
    Uliana Vinogradova\thanks{SB Robotics, Russian Federation}
}

\date{}

\begin{document}

\maketitle

\begin{abstract}%
    We suggest a novel procedure for online change point detection. Our approach expands an idea of maximizing a discrepancy measure between points from pre-change and post-change distributions. This leads to flexible algorithms suitable for both parametric and nonparametric scenarios. We prove non-asymptotic bounds on the average running length of the procedure and its expected detection delay. The efficiency of the algorithm is illustrated with numerical experiments on synthetic and real-world data sets.
\end{abstract}

\section{Introduction}

The problem of change point detection is familiar to statisticians and machine learners since the pioneering works of \citet{page54, page55}, \citet{shiryaev61, shiryaev63} and \citet{roberts66} but, nevertheless, it
still attracts attention of many researchers due to its practical importance. In our paper, we assume that a learner observes independent random elements $X_1, \dots, X_t, \dots$ arriving successively. There exists a moment $\tstar \in \mathbb N$ (not accessible to the statistician), such that $X_1, \ldots, X_\tstar$ are drawn from a distribution, which has a density $\sfp$ with respect to a dominating measure $\m$, while $X_{\tstar + 1}, \dots, X_t, \dots$ have a density $\sfq$ (with respect to the same measure), which differs from $\sfp$. The measure $\m$ is not restricted to be the Lebesgue measure, it can be equal to the counting measure (in the discrete case) or the Hausdorff measure on a low-dimensional manifold as well. The learner is interested in reporting about the occurrence of $\tstar$ as fast as possible while keeping the false alarm rate at an acceptable level. This problem is called online (also referred to as sequential or quickest) change point detection. Such a setup is quite different from another major research direction, offline change point detection \citep{ds01, zou14, mj14, ds15, biau16, korkas17, ga18, arlot19, padilla21, corradin22, malte22}, where the statistician has an access to the whole time series at once, and, instead of taking decisions on the fly, they are mostly interested in a retrospective analysis and change point localization.

The complexity of a change point detection problem severely depends on the data generating mechanism. The most popular one is a mean shift, that is, $\E X_{\tstar} \neq \E X_{\tstar + 1}$. Plenty of papers are devoted to a mean shift detection in a univariate or multivariate Gaussian sequence (see, for instance, \citep{sugiyama08, kanamori09, eh13, pein17, rinaldo21, chen22, sun22}), but the recent research \citep{eichinger18, maillard2019sequential, wang20, yu20a, yu20b} also considers a more general sub-Gaussian noise. One usually exploits CUSUM-type or likelihood-ratio-type test statistics to perform this task. A broader problem of parametric change point detection (see, for example, \citep{cao18, dette20, yu20a, corradin22, sun22, titsias22}) admits that $\sfp$ and $\sfq$ belong to a parametric family of densities $\cp = \{\sfp_\theta : \theta \in \Theta\}$. In this setup, the distribution change detection is reduced to detection of a shift in the underlying parameter $\theta \in \Theta$. A special case of parametric setup includes changes in covariance and correlation coefficients \citep{bolognani13, chaudhuri21, chaudhuri2024round, chaudhuri22}.
Both the mean shift model and the parametric change point detection require strong modelling assumptions which are likely to be violated in practical applications. In our paper, we are mostly interested in a nonparametric change point detection problem \citep{hero06, hbm08, zou14, li15, biau16, ga18, arlot19, kurt21, padilla21b, shin22, ferrari23}. We do not impose restrictive conditions on the densities $\sfp$ and $\sfq$. However, the procedure we propose is quite universal in a sense that it is suitable for different setups, including, for instance, the nonparametric one and the mean shift detection in a multivariate Gaussian sequence model.

Though the number of papers on change point detection is huge and many of them are devoted to theoretical analysis of the procedures (see, e.g., \citep{pollak09, tartakovsky12, li15, cao18, yu20b, liang21, chen22, chu22, dehling22, shin22}), nonparametric change point detection is studied not so well. Some papers provide rigorous guarantees on the average running length of the procedures (i.e. the expected number of iterations the algorithm makes in a stationary regime\footnote{Here and further in this paper, the stationary regime corresponds to the situation when the elements of the sequence $\{X_t : 1 \leq t \leq T\}$ have the same distribution.} until a false alarm) but, to our knowledge, there are no non-asymptotic high probability bounds on the detection delay.

Let us describe the idea of our algorithm. In the sequential change point detection, at the moment $t$, one usually tests the hypothesis
\begin{equation}
    \label{eq:h0}
    H_0 : \text{ $X_1, \dots, X_t$ have the same distribution}
\end{equation}
against the composite alternative
\begin{equation}
    \label{eq:h1}
    H_1 : \text{there exists $\tau \in \{1, \dots, t - 1\}$, such that $\tstar = \tau$,}
\end{equation}
which can be considered as the union of the alternatives of the form $H_1^\tau : \tstar = \tau$, $\tau \in \{1, \dots, t - 1\}$.
If the change occurred at some $\tau \in \{1, \dots, t-1\}$ (that is, $H_1^\tau$ takes place), then the distribution of $X_1, \dots, X_\tau$ must differ from the one of $X_{\tau + 1}, \dots, X_t$. To detect such a discrepancy, we introduce an auxiliary function $D : \sfx \rightarrow (0, 1)$ that should distinguish between the pre-change and post-change distributions. The higher values of $D(X)$ reflect a larger confidence that $X$ was drawn from the density $\sfp$, rather than from $\sfq$. Such an approach of reducing an unsupervised learning problem to a supervised one is not new (see, e.g., \cite[Section 14.2.4]{htf09}) and was used in the problems of density estimation \citep{gutmann12}, generative modelling \citep{goodfellow14, grover19}, and density ratio estimation \citep{grover19}.
Based on this idea, \citeauthor*{hushchyn20} designed an algorithm for change point detection. However, the sliding window technique the authors used leads to significant detection delays. Besides, \citeauthor*{hushchyn20} do not provide any theoretical guarantees on the running length and the detection delay of their procedure.

Let us fix $t \in \mathbb N$ and a change point candidate $\tau \in \{1, \dots, t-1\}$. In order to find a good auxiliary classifier $D$, distinguishing between $X_1, \dots, X_\tau$ and $X_{\tau + 1}, \dots, X_t$, we fix a family $\cd$ of functions taking their values in $(0, 1)$ and choose a maximizer of the cross-entropy
\begin{equation}
    \label{eq:discrepancy}
    \frac{\tau (t - \tau)}t
    \Bigg[ \frac1\tau \sum\limits_{s=1}^\tau \log\big( 2D(X_s) \big)
    + \frac1{t - \tau} \sum\limits_{s = \tau + 1}^t \log\big(2 - 2D(X_s) \big) \Bigg]
\end{equation}
over $\cd$. A similar approach was introduced in \citep{gutmann12, goodfellow14} but for the purposes of density estimation and generative modelling, respectively. In the context of sequential change point detection, \citet*{li15}, as well as \citet*{chang18}, used a different divergence measure, the squared maximum mean discrepancy, to derive a kernel change point detection method. In our paper, we adapt the technique of \citep{goodfellow14} for the quickest change point detection. Following \citep{gutmann12, goodfellow14}, we call our approach \emph{noise-contrastive} and refer to the function $D$ as discriminator.

We show in Section \ref{sec:nc_approach} that our algorithm needs to approximate $\log(\sfp/\sfq)$ with a reasonable accuracy to be sensitive to distribution changes. This makes it similar to change point detection methods based on the density ratio estimation \citep{liu13, hushchyn20, hushchyn21}. For instance, \citeauthor*{liu13} use KLIEP \citep{sugiyama08}, uLSIF \citep{kanamori09} and RuLSIF \citep{yamada13} for online change point detection. In \citep{hushchyn21}, the authors use the $\alpha$-relative chi-squared divergence, the same functional as in RuLSIF \citep{yamada13}, to construct a change point detection procedure. The advantage of such methods is that the estimation of the ratio $\sfp/\sfq$ can be a much easier task than estimation of the densities $\sfp$ and $\sfq$ themselves. However, in the density-ratio based algorithms the authors usually use a sliding window technique and compare the distributions between two large non-overlapping segments of the time series. This approach shows a good performance in the offline setup, when the learner is interested in  change point estimation, but leads to large detection delays in the online case.  In our paper, we adjust the test statistic in order to make it suitable for the sequential detection problem. 
Besides, in contrast to \citep{liu13, hushchyn20, hushchyn21}, we study the detection delay of our procedure and the behaviour of the test statistic under the null hypothesis.

\medskip

\noindent{\bfseries Contribution.} \quad
We suggest an algorithm for sequential change point detection based on the noise-contrastive approach and online convex optimization (Algorithm \ref{alg:fast_change_point}).
We provide non-asymptotic large deviation bounds on its running length and detection delay (Theorem \ref{th:rl_dd_falcon}) and discuss their optimality. 
Algorithm \ref{alg:fast_change_point} shows promising results in numerical experiments on synthetic and real-world data sets, outperforming strong competitors.

\medskip

\noindent{\bfseries Organization of the paper.}\quad
The rest of the paper is organized as follows.
In Section \ref{sec:alg_description}, we elaborate on noise-contrastive approach to change point detection and introduce our algorithm (Algorithm \ref{alg:fast_change_point}), based on tools from online convex optimization.
In Section \ref{sec:theoretical}, we derive non-asymptotic large deviation bounds on the running length and the detection delay of our procedure (Theorem \ref{th:rl_dd_falcon}).
Section \ref{sec:numerical} is devoted to numerical experiments.
Section \ref{sec:lemmata_proofs} collects the proofs of our main results, presented in Sections \ref{sec:alg_description} and \ref{sec:theoretical}. 
Some auxiliary results are deferred to appendices.

\medskip

\noindent{\bfseries Notation.}\quad
We use the following notations throughout the paper.
For $s \ge 1$ and a probability density $\sfp$, we define the $L_s(\sfp)$-norm as $\|f\|_{L_s(\sfp)} = \left(\E_{\xi \sim \sfp} |f(\xi)|^s\right)^{1/s}$.
Given two probability measures with the densities $\sfp \ll \sfq$, $\KL(\sfp, \sfq) = \int \sfp(x) \log(\sfp(x)/\sfq(x)) \dd \m$ stands for
the Kullback-Leibler divergence between $\sfp$ and $\sfq$. For any two densities $\sfp$ and $\sfq$,
\[
    \JS(\sfp, \sfq) = \frac12 \KL\left(\sfp, \frac{\sfp + \sfq}2 \right) + \frac12 \KL\left(\sfq, \frac{\sfp + \sfq}2 \right)
\]
denotes the Jensen-Shannon divergence between $\sfp$ and $\sfq$.

\section{Algorithm description}
\label{sec:alg_description}

This section aims to provide a detailed description of our algorithm. We start with some intuition behind the procedure. Then we briefly introduce the online convex optimization framework and show how it applies to sequential change point detection. Finally, we present our method in Algorithm \ref{alg:fast_change_point}. 

\subsection{Noise-contrastive approach}
\label{sec:nc_approach}

The main idea of the noise-contrastive approach is to maximize the discrepancy measure \eqref{eq:discrepancy} for each change point candidate $\tau \in \{1, \dots, t-1\}$. Since the classifier $D$ in \eqref{eq:discrepancy} must take its values in $(0, 1)$ we consider a parametric class
\begin{equation}
    \label{eq:d_parametrization}
    \left\{ D_\theta(x) = e^{\theta^\top \psi(x)} / (1 + e^{\theta^\top \psi(x)}): \theta \in \Theta \right\},
\end{equation}
where $\Theta \subset \R^d$ is a compact convex set and $\psi : x \mapsto \big(\psi_1(x), \dots \psi_d(x) \big)^\top$ is a fixed vector-function. Applying parametrization \eqref{eq:d_parametrization} to the discrepancy \eqref{eq:discrepancy}, we obtain a statistic
\begin{align}
    \label{eq:t}
    \ct_{\tau, t}(\theta)
    = \frac{t - \tau}t \sum\limits_{s=1}^\tau \left[ \theta^\top \psi(X_s) - \log\left( \frac{1 + e^{\theta^\top \psi(X_s)}}2 \right) \right]
    - \frac{\tau}t \sum\limits_{s = \tau + 1}^t \log\left( \frac{1 + e^{\theta^\top \psi(X_s)}}2 \right),
\end{align}
where $\theta \in \Theta$.
The cornerstone of our approach is the basic property of $\ct_{\tau, t}(\theta)$ formulated in the following lemma.

\begin{Lem}
    \label{lem:js}
    Let $t \in \mathbb N$. Assume that the change point occurred at some $\tstar \in \{1, \dots, t - 1\}$.
    Then it holds that
    \begin{equation}
        \label{eq:t_exp}
        \E \ct_{\tstar, t}(\theta)
        \geq \frac{2\tstar(t - \tstar)}t \left( \JS(\sfp, \sfq) - \frac1{8} \left\|\theta^\top \psi - \log(\sfp / \sfq) \right\|^2_{L_2((\sfp + \sfq) / 2 )} \right)
        \quad \text{for all $\theta \in \Theta$.}
    \end{equation}
\end{Lem}

Lemma \ref{lem:js} suggests that we have to choose the components of $\psi$ properly (e.g., Legendre or Hermite polynomials, splines, wavelets, etc.) to ensure that the class $\{\theta^\top \psi(x) : \theta \in \Theta\}$ approximates $\log(\sfp/\sfq)$ with a reasonable accuracy and that the right-hand side of \eqref{eq:t_exp} is positive for some $\theta \in \Theta$. This makes our procedure similar to the change point detection methods based on density ratio estimation \citep{liu13}. If the class $\{\theta^\top \psi(x) : \theta \in \Theta\}$ is rich enough, then the expectation of
\begin{align}
    \label{eq:s}
    \cs_t = \max_{1 \leq \tau \leq t - 1} \max_{\theta \in \Theta} \ct_{\tau, t}(\theta)
\end{align}
starts to grow once a change point occurred.
On the other hand, if $X_1, \dots, X_t$ are i.i.d. random elements, it is easy to check that $\E \ct_{\tau, t}(\theta) \leq 0$ for all $\tau \in \{1, \dots, t - 1\}$ and $\theta \in \Theta$. In this case, we should expect mild values of the statistic $\cs_t$. This makes $\cs_t$ a good candidate for the discrepancy measure between pre-change and post-change samples. We illustrate this point with a simple example.

\begin{figure}[ht]
    \centering
    \subfloat{
        \includegraphics[width=0.48\textwidth]{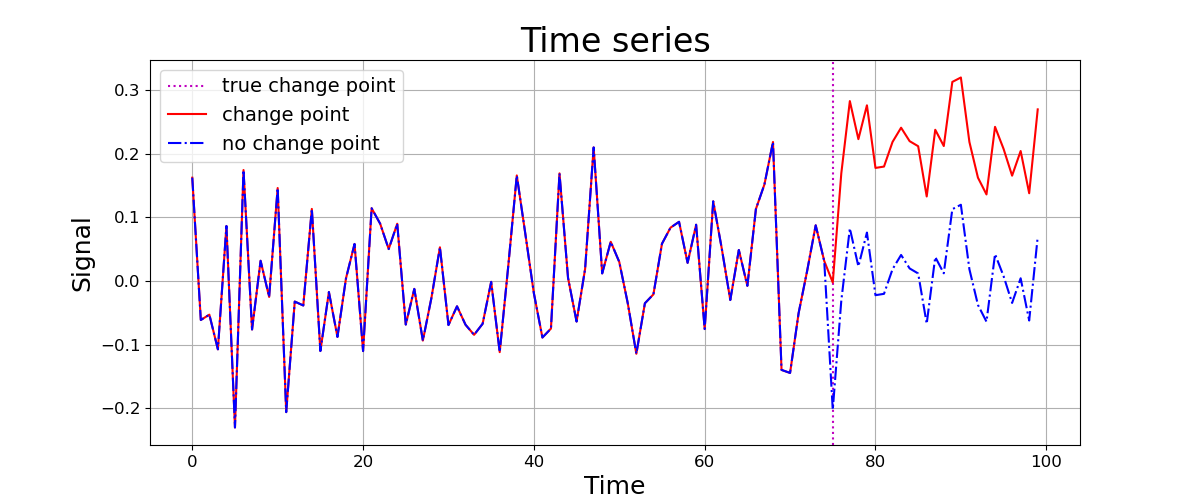}
    }
    \hfill
    \subfloat{
        \includegraphics[width=0.44\textwidth, height=3cm]{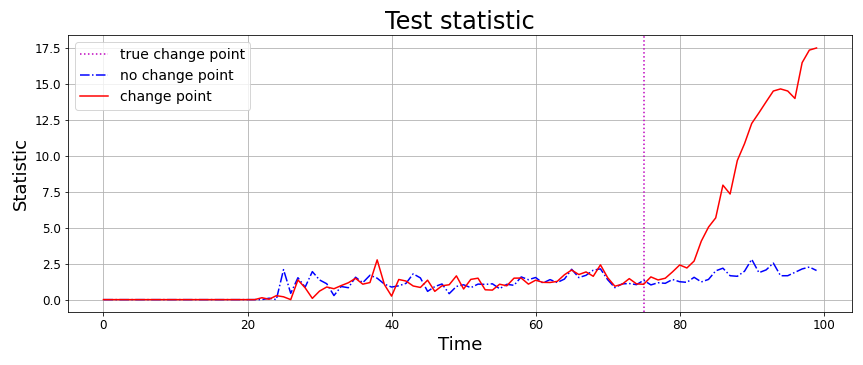}
    }
    \caption{An example of behaviour of the statistic $\cs_t$ (defined in \eqref{eq:s}) in the presence of a change point and in the stationary regime. Left: a stationary sequence (blue) and a sequence of observations with a change point (red). Right: corresponding values of the test statistic $\cs_t$. The dashed vertical line corresponds to the change point $\tstar$.}
    \label{fig:illustration}
\end{figure}

Let $X_1, \dots, X_T$ with $T = 100$ be a sequence of i.i.d. observations drawn according to the Gaussian distribution $\mathcal N(0, 0.01)$. Let $\tstar = 75$ and define a sequence $Y_1, \dots, Y_T$ according to the formula
\[
    Y_t =
    \begin{cases}
        X_t, \quad \text{if $t \leq \tstar$},\\
        0.2 + X_t, \quad \text{otherwise}.
    \end{cases}
\]
In other words, the sequences $\{X_t : 1 \leq t \leq T\}$ and $\{Y_t : 1 \leq t \leq T\}$ coincide before the change point $\tstar$ and differ by the shift equal to $0.2$ after it.
A realization of the sequences is displayed in Figure \ref{fig:illustration}.
Since the density ratio $\log(\sfp(x) / \sfq(x)) = -20x + 2$ is an affine function in this setup, it suffices to set $\psi(x) = (1, x)^\top$ and $\Theta = \cb(0, 25)$ to ensure that the approximation error is zero. 
We observe that the statistic $\cs_t$, computed for the sequence $Y_1, \dots, Y_T$ (the solid red line in Figure \ref{fig:illustration}), increases sharply after the change point (see Figure \ref{fig:illustration}, vertical line) grows to the value of about $17.5$ by the end of the sequence. 
In contrast, it never exceeds $2.5$ in the stationary regime (see the dotted blue line in Figure \ref{fig:illustration}).

The only drawback of the statistic \eqref{eq:s} is that it requires to maximize $\ct_{\tau, t}(\theta)$, $1 \leq \tau \leq t - 1$, over $\theta \in \Theta$ on each round from scratch. As a result, the total computational cost becomes prohibitive for real-world tasks. For this reason, we develop a strategy for approximate computation of $\cs_t$ based on online convex optimization. It allows us to update the approximate maximizer of $\ct_{\tau, t}(\theta)$ over $\theta \in \Theta$ in a recursive manner, using the results from the previous iterations $\{1, \dots, t - 1\}$. This leads to a significant speedup of the procedure.

\subsection{Online convex optimization}
\label{sec:oco}

Our approach relies on the tools from online convex optimization, so let us recall its framework to the reader before we move to the description of the algorithm.
We also refer to the brilliant surveys of \citet{shalev-shwartz12} and \citet{hazan16} for the introduction and basic algorithms on this topic.
Online convex optimization (OCO) is a repeating game between a learner (also referred to as player) and his opponent (or adversary).
On the round $t \in \{1, \dots, T\}$, the player chooses a prediction $\widehat p_{t - 1}$ from a given convex compact set $\mathfrak P$.
After that, the opponent reveals a convex function $\ell_t : \mathfrak P \rightarrow \R$, and the learner suffers the loss $\ell_t(\widehat p_{t - 1})$.
Finally, the player updates its prediction using an algorithm $\ca$.
Its performance is measured by the regret after $T$ rounds, defined as
\begin{align*}
    \regret_{\ca}(T) = \sum_{t = 1}^T \ell_t(\widehat p_{t - 1}) - \min_{p \in \mathfrak P} \sum_{t = 1}^T \ell_t(p) .
\end{align*}
The goal of the player is to make $\regret_\ca(T)$ as small as possible regardless the opponent's strategy.
We summarize the described framework below.

\medskip
\begin{tcolorbox}
    \textbf{Online convex optimization framework.}
    \begin{itemize}
        \item A convex compact set $\mathfrak P$ is given.
        \item \textbf{For} $t = 1, 2, \dots, T, \dots $:
        \begin{enumerate}
            \item the learner makes a prediction $\widehat p_{t - 1}$;
            \item the adversary reveals a convex function $\ell_t$, and the learner suffers the loss $\ell_t(\widehat p_{t - 1})$;
            \item the learner calculates $\widehat p_t = \ca(\ell_1, \dots, \ell_t) \in \mathfrak P$.
        \end{enumerate}
    \end{itemize}
\end{tcolorbox}

In the context of sequential change point detection, the online convex optimization framework was applied in the paper of \citet*{cao18}. However, the authors imposed strong parametric assumptions on the density of observations. In particular, the pre-change and post-change densities, $\sfp$ and $\sfq$ respectively, must belong to an exponential family with known link function.
In our approach, we relax such assumptions, because we make an assumption about the class $\{\theta^\top \psi(x) : \theta \in \Theta\}$, which we are free to choose.

\subsection{The algorithm}

Let us describe how the OCO framework applies to the noise-contrastive approach for sequential change point detection.
Note that, for any $t \in \mathbb N$, any $\tau \in \{1, \dots, t-1\}$, and any $\theta \in \Theta$, the statistic $\ct_{\tau, t}(\theta)$ satisfies the equality 
\begin{align*}
    t \ct_{\tau, t}(\theta) - (t - 1) \ct_{\tau, t-1}(\theta)
    = - \sum\limits_{s = 1}^\tau \log\left( \frac{1 + e^{-\theta^\top \psi(X_s)}}2 \right)
    - \tau \log\left( \frac{1 + e^{\theta^\top \psi(X_t)}}2 \right).
\end{align*}
With the convention $\ct_{\tau, t}(\theta) = 0$ for any $\tau \geq t$ and $\theta \in \Theta$, we can express $t \, \ct_{\tau, t}(\theta)$ recursively in the following form:
\[
    -t \, \ct_{\tau, t}(\theta) = -(t - 1) \, \ct_{\tau, t - 1}(\theta) + \tau \varphi_{\tau, t}(\theta),  
\]
where
\begin{equation}
    \label{eq:phi}
    \varphi_{\tau, t}(\theta) =
    \begin{cases}
        \frac1\tau \sum\limits_{s = 1}^\tau \log\left( 1 + e^{-\theta^\top \psi(X_s)} \right)
        + \log\left(1 + e^{\theta^\top \psi(X_t)} \right) - 2 \log 2, \quad \text{if $\tau \leq t - 1$,}\\
        0, \quad \text{otherwise}.
    \end{cases}
\end{equation}
Let us fix $\tau \in \mathbb N$ and run the online convex optimization game with the domain $\mathfrak P = \Theta$ and the loss $\ell_t(\theta) = \varphi_{\tau, t}(\theta)$ (which is convex). Assume that the player made predictions $\widehat \theta_{\tau, 0}, \dots, \widehat \theta_{\tau, t - 1}$
according to an OCO algorithm $\ca$ on the first $t$ rounds, and consider
\[
    \widehat \ct_{\tau, t} = -\frac{\tau}t \sum\limits_{s = 1}^t \varphi_{\tau, s}(\widehat \theta_{\tau, s}) = \frac{t-1}t \widehat \ct_{\tau, t-1} -\frac{\tau}t \varphi_{\tau, t}(\widehat \theta_{\tau, t}).
\]
The difference between $\widehat \ct_{\tau, t}$ and the maximum of $\ct_{\tau, t}(\theta)$ is proportional to the per-round regret of $\ca$:
\[
    -\widehat\ct_{\tau, t} + \max\limits_{\theta \in \Theta} \ct_{\tau, t}(\theta)
    = \frac{\tau}{t}\sum_{s = 1}^t \left( \varphi_{\tau, s}(\widehat\theta_{\tau, s - 1}) - \min\limits_{\theta \in \Theta} \sum_{s = 1}^t \varphi_{\tau, s}(\theta) \right)
    =
    \begin{cases}
        \tau \, \regret_\ca(t) / t,
        \quad \text{if $t > \tau$,}\\
        0 \quad \text{otherwise.}
    \end{cases}
\]
It the regret of $\ca$ is sublinear, then, for any $\tau \in \mathbb N$, the value of $\widehat \ct_{\tau, t}$ will approach to the maximum of $\ct_{\tau, t}(\theta)$ over $\theta \in \Theta$ as $t$ tends to infinity. 
Hence, instead of the statistic $\cs_t$, defined in \eqref{eq:s}, we can use  $\widehat \cs_{t} = \max\limits_{1 \leq \tau \leq t - 1} \widehat \ct_{\tau, t}$. This brings us to the following algorithm.

\medskip

\begin{tcolorbox}[breakable, enhanced]
    \begin{Alg}[FALCON, \underline{f}ast \underline{al}gorithm based on \underline{con}trastive approach]
    \hfill
    \label{alg:fast_change_point}
    \begin{itemize}
        \item \textbf{Input:} an OCO algorithm $\ca$, a decision domain $\Theta$, and a threshold $\z > 0$.
        \item \textbf{Initialization:} $\widehat \theta_{\tau, t} = 0$ and $\widehat \ct_{\tau, t} = 0$ for all $\tau \geq t$ such that $\tau \in \N$ and $t \in \mathbb N \cup \{0\}$.
        \item \textbf{For} $t = 1, 2, \dots$ \textbf{do} the following.
        \begin{enumerate}
            \item Receive an observation $X_t$.
            \item For each $\tau \in \{1, \dots, t - 1\}$, compute
            \[
                \widehat \ct_{\tau, t}
                = -\frac{\tau}t \sum\limits_{s = 1}^t \varphi_{\tau, s}(\widehat \theta_{\tau, s - 1})
                = \frac{t-1}t \, \widehat \ct_{\tau, t-1} - \frac{\tau}t \, \varphi_{\tau, t}(\widehat \theta_{\tau, t - 1}),
            \]
            where the function $\varphi_{\tau, t}(\theta)$ is defined in \eqref{eq:phi}.
            \item Compute the test statistic
            \[
                \widehat \cs_{t} = \max\limits_{1 \leq \tau \leq t - 1} \widehat \ct_{\tau, t},
            \]
            \item If $\widehat \cs_t > \z$, terminate the procedure, report the change point occurrence, and return the stopping time $\widehat t$. Otherwise, update the estimates $\widehat \theta_{\tau, t}$, $1 \leq \tau \leq t - 1$, according to the online learning algorithm:
            \begin{equation}
                \label{eq:theta_update}
                \widehat \theta_{\tau, t} = \ca(\varphi_{\tau, 1}, \dots, \varphi_{\tau, t}) \in \Theta
            \end{equation}
        \end{enumerate}
        \item \textbf{Return}.
    \end{itemize}
    \end{Alg}
\end{tcolorbox}

\bigskip

Though, in general, Algorithm \ref{alg:fast_change_point}, admits an arbitrary online convex optimization procedure as a subroutine, we recommend practitioners to take the following fact into account.

\begin{Lem}
    \label{lem:exp-concave}
    Let $|\theta^\top \psi(x)| \leq B$ for all $\theta \in \Theta$ and almost all $x$.
    Then, for any $t \in \mathbb N$, any $1 \leq \tau \leq t - 1$, and any $\alpha \leq 0.5 e^{-B}$, the function $\varphi_{\tau, t}$, defined in \eqref{eq:phi}, is $\alpha$-exp-concave\footnote{Recall that a function $f : \Theta \to \R$ is called $\alpha$-\emph{exp-concave} if $\exp\{-\alpha f(\theta) \}$ is concave on $\Theta$.} on the convex set $\Theta$. 
\end{Lem}
The proof of Lemma \ref{lem:exp-concave} is deferred to Appendix \ref{sec:lem_exp-concave_proof}. OCO algorithms, which are able to leverage the loss curvature, were discussed, for instance, in \cite{hazan07}. In particular, \citet*{hazan07} showed that the Follow the Approximate Leader (FTAL) and Online Newton Step (ONS) strategies achieve logarithmic regret in online exp-concave optimization. For this reason, we find them appropriate for our purposes. One can also use modifications of these algorithms, such as LightONS \citep{wang26}.

The running time of Algorithm \ref{alg:fast_change_point} depends on the subroutine $\ca$. If $\ca$ requires $c_\ca$ operations to update the prediction, then the total computational cost of Algorithm \ref{alg:fast_change_point} after $t$ iterations will be $\co(t^2 c_\ca)$. For the FTAL and ONS algorithms $c_\ca$ is $\co(d^3)$, where $d$ is the dimension of $\theta$. We would like to emphasize that this is much smaller than a time needed to maximize $\ct_{\tau, t}(\theta)$ within a reasonable accuracy. As a consequence, one iteration of Algorithm \ref{alg:fast_change_point} is significantly faster than naive computation of the statistic $\cs_t$ given by \eqref{eq:s}.
Moreover, there is a simple way to reduce computational time of Algorithm \ref{alg:fast_change_point} to $\co(t \tau_{\min} c_\ca)$, where $\tau_{\min}$ is given by \eqref{eq:tau_min}. One just needs to replace $\widehat \cs_t$ by
\begin{equation}
    \label{eq:tilde_s}
    \widetilde\cs_t = \max\limits_{t - 2 \tau_{\min} \leq \tau \leq t} \widehat\ct_{\tau, t}.
\end{equation}

In Section \ref{sec:theoretical}, we discuss that this modification does not affect the ability of Algorithm \ref{alg:fast_change_point} to detect change points.

\section{Theoretical properties}
\label{sec:theoretical}

Lemma \ref{lem:js} may give some intuition why the test statistic $\widehat\cs_t$ is appropriate for change point detection. However, it still does not provide a quantitative upper bound on the detection delay of Algorithm \ref{alg:fast_change_point} and, more importantly, does not discuss behaviour of the test statistic in the stationary regime (that is, when $\tstar = \infty$ and then $X_1, \dots, X_T$ are i.i.d.). The latter is crucial for a proper choice of the threshold $\z$ and a rigorous study of the running length of the procedure. This section aims to fill these gaps. We start with a preliminary but insightful upper bound on the statistic $\widehat\ct_{\tau, t}$ for fixed $t \in \{1, \dots, T\}$ and $\tau \in \{1, \dots,  t - 1\}$.

\begin{Th}
    \label{th:ftal_test_stat_upper_bound}
    Let us fix arbitrary positive integers $\tau$ and $t$ such that $\tau < t$, and assume that $X_1, \dots, X_t \sim \sfp$ are i.i.d. random elements.
    Assume further that $|\theta^\top \psi(X)| \leq B$ for all $\theta \in \Theta$ and almost all $X \sim \sfp$.
    Then, for any $\delta \in (0, 1)$, with probability at least $(1 - \delta)$, it holds that
    \begin{align}
    \label{eq:null_upper_bound}
        \widehat\ct_{\tau, t}
        \leq \frac{3 e^B d}{\tau}
        + \frac{19 B \log(4 / \delta)}{4 \tau}
        + \frac{31 e^B \log(4 / \delta)}{6 \tau}.
    \end{align}
\end{Th}

The proof of Theorem \ref{th:ftal_test_stat_upper_bound} is deferred to Appendix \ref{sec:th_ftal_test_stat_upper_bound_proof}, and it has two challenges. First, the loss $\varphi_{\tau, t}(\theta)$ depends not only on $X_t$ but also on $X_1, \dots, X_\tau$. This makes a popular approach for analysis of online algorithms, based on martingale concentration inequalities, inapplicable in our case. We suggest a framework combining Bernstein's  inequality for martingales \citep{freedman75} and localization technique for empirical processes \citep{bbm05}. It might be informative for specialists in statistical learning, because, to our knowledge, there was no need to use such a combination in other learning problems. Second, a naive variance bound in the martingale Bernstein inequality by a constant will lead to non-optimal guarantees. In order to get a sharp result, we need to use a bit more sophisticated technique.

\begin{Rem}
\label{rem:st_upper_bound}
Based on Theorem \ref{th:ftal_test_stat_upper_bound}, we can set
\begin{equation}
    \label{eq:falcon_threshold}
    \z = 3e^B d + \frac{19B}{4} \log \big( 2T (T - 1) / \delta \big) + \frac{31 e^B}{6} \log \big( 2T(T - 1) / \delta \big).
\end{equation}
Then Theorem \ref{th:ftal_test_stat_upper_bound} and the union bound imply that, with probability at least $(1 - \delta)$, Algorithm \ref{alg:fast_change_point} has the running length at least $T$:
\[
    \max\limits_{1 \leq t \leq T} \widehat\cs_t = \max\limits_{1 \leq t \leq T} \max\limits_{1 \leq \tau \leq t - 1} \widehat\ct_{\tau, t}
    \leq 3e^B d + \frac{19B}{4} \log\big(2T (T - 1) / \delta \big) + \frac{31 e^B}{6} \log\big( 2T(T - 1) / \delta \big)
    = \z.
\]
\end{Rem}

Theorem \ref{th:ftal_test_stat_upper_bound} also indicates a way for further improvement of Algorithm \ref{alg:fast_change_point}. If we know in advance that $\tstar \geq \tau_0$ (i.e., there is a warm-up period where we can collect some data), then we can use a test statistic $\cs_t^\circ = \max \big\{ \widehat\ct_{\tau, t} : \tau_0 \leq \tau \leq t - 1 \big\}$, instead of $\widehat \cs_t$. 
In this case, Theorem \ref{th:ftal_test_stat_upper_bound} and the same argument as in Remark \ref{rem:st_upper_bound} provide a bound, that is $\tau_0$ times sharper, thereby allowing for a smaller threshold $\z$ without decrease in the running length. This results in a sharper bound on the detection delay and distinguishes our approach from methods based on the sliding window technique (e.g., \cite{liu13, li15, hushchyn20}), where test statistics are typically constructed from a limited number of observations within the window and do not account for the earlier reference period.

We proceed with a high-probability upper bound on the  detection delay of Algorithm \ref{alg:fast_change_point}. The main result of this paper is summarized in the following theorem.

\begin{Th}
\label{th:rl_dd_falcon}
    Assume that $|\theta^\top \psi(X)| \leq B$ for all $\theta \in \Theta$ almost surely on the supports of $\sfp$ and $\sfq$.
    For any $t \in \N$, let $\regret_{\ca}(t)$ stand for the regret of the online convex optimization algorithm $\ca$ applied to the functions $\{ \varphi_{\tstar, s} \, : \, 1 \leq s \leq t \}$ after $t$ rounds, and let  $\mar_{\tstar} = \max\{\regret_\ca(t) / t : t \geq \tstar\}$ denote the maximum average regret.
    Let us introduce
    \begin{align}
    \label{eq:rho_theta}
        \rho(\Theta) = \min_{\theta \in \Theta} \big\|\theta^\top \psi - \log(\sfp / \sfq)\big\|_{L_2((\sfp + \sfq) / 2)} .
    \end{align}
    Take an arbitrary $\delta \in (0, 1)$ and set the threshold $\z$ as specified in \eqref{eq:falcon_threshold}.
    Then the following holds:
    \begin{itemize}
        \item if $\tstar = \infty$, then Algorithm \ref{alg:fast_change_point} makes at least $T$ steps until the false alarm with probability at least $(1 - \delta)$;
        \item otherwise, if $\tstar$ is sufficiently large in the sense that it satisfies
        \begin{equation}
            \label{eq:tau_min}
            \tstar
            \geq \tau_{\min} = 2 \ceil{\frac{\z + \tstar \, \mar_{\tstar} + (3B + 8)\log(1 / \delta)}{\JS(\sfp, \sfq) - \rho^2(\Theta) / 2}},
        \end{equation}
        then the stopping time $\widehat{t}$ of Algorithm \ref{alg:fast_change_point} fulfills $\widehat t - \tstar \leq \tau_{\min}$, with probability at least $(1 - \delta)$.
    \end{itemize}
\end{Th}

We postpone the proof of Theorem \ref{th:rl_dd_falcon} to Appendix \ref{sec:th_rl_dd_falcon_proof}. Let us note that, according to this theorem, Algorithm \ref{alg:fast_change_point} detects a change point in $\tau_{\min}$ (see \eqref{eq:tau_min}) iterations with high probability. This means that there is no need to compute $\widehat\ct_{\tau, t}$ for clearly poor change-point candidates (for example, $\tau \leq t - 2 \tau_{\min})$. Instead, we can use the statistic $\widetilde\cs_t$ from \eqref{eq:tilde_s}, significantly reducing the required computational resources.

\begin{Rem} 
    \label{rem:ons_ftal_dd}
    If we assume that $\Theta \subseteq \cb(0, b) \subset \R^d$ and $\|\psi(X)\| \leq R$ almost surely, then $\varphi_{\tau, t}(\theta)$ is exp-concave on $\Theta$ and, moreover, the norm of its gradient does not exceed
    \[
        \left\| \nabla \varphi_{\tau, t}(\theta) \right\| \leq \frac{1}{\tau} \sum_{s = 1}^\tau \left\| \nabla \log\left(1 + e^{-\theta^\top \psi(X_s)} \right) \right\| + \left\| \nabla \log\left(1 + e^{\theta^\top \psi(X_t)}\right) \right\|
        \leq 2bR
    \]
    for all $\theta \in \Theta$. According to \citet[Theorems 2 and 6]{hazan07}, the regrets of ONS and FTAL (with properly tuned parameters) are not greater, than
    \[
        \regret_{\mathrm{ONS}}(t)
        \leq 5d (2 e^{bR} + 4bR) \log t
        \quad \text{and} \quad
        \regret_{\mathrm{FTAL}}(t)
        \leq 64 d( 2e^{bR} + 4bR)(1 + \log t).
    \]
    The corresponding values of $\tstar \, \mar_{\tstar}$ are bounded by $5d (2 e^{bR} + 4bR) \log \tstar$ and $64 d( 2e^{bR} + 4bR)(1 + \log \tstar)$, respectively.
    Therefore, in both cases we obtain the detection delay bound
    \begin{align*}
        \widehat t - \tstar = \co\left( \frac{d e^{bR} \log(T \tstar / \delta)}{\JS(\sfp, \sfq) - \rho^2(\Theta) / 2} \right)
    \end{align*}
    on an event of probability at least $(1 - \delta)$.
\end{Rem}

As it becomes clear from Remark \ref{rem:ons_ftal_dd}, Theorem \ref{th:rl_dd_falcon} provides a quantitative characterization of the trade-off between the richness of the class $\{\theta^\top \psi(x) : \theta \in \Theta\}$ and its complexity.
Larger set $\Theta$ and dimension $d$ improve the approximation accuracy of $\log(\sfp / \sfq)$ and reduce $\rho(\Theta)$. Conversely, the numerator in the detection delay bound increases with the dimension and the size of the set $\Theta$.
If we assume that $\log(\sfp / \sfq)$ belongs to a H\"older class $\ch^\beta(\sfx)$, where $\sfx \subseteq [-1, 1]^k$, then we can approximate it within the accuracy $\sqrt{\JS(\sfp / \sfq)}$ with respect to the $L_2\big( (\sfp + \sfq) / 2 \big)$, using, for instance, a polynomial of degree $d = \co\big(\JS(\sfp, \sfq)^{-k/(2\beta)}\big)$. In this case, $\widehat t - \tstar = \co\big( \JS(\sfp, \sfq)^{-(2\beta + k)/(2\beta)} \big)$, which corresponds to the minimax optimal rates of density estimation with respect to the Jensen-Shannon divergence (see, e.g., \cite{puchkin24}).

The result of Theorem \ref{th:rl_dd_falcon} can be easily extended to the case of multiple change points. We just have to restart the procedure each time a change in distribution was detected. The only additional requirement will be that the distance between two subsequent change points is at least $\Omega(\log T)$, which is quite standard for the offline setup (see, e.g., \cite[Assumption 3]{yu20a} and \cite[Assumption 2]{wang20}).
We also emphasize that $\widehat t$  is the stopping time of the procedure, it should not be confused with an estimate of $\tstar$. In the present paper, we focus on the running length and the detection delay only. We do not tackle the problem of localization of $\tstar$, which is usually considered in \emph{offline} change point detection.

\section{Numerical Experiments}
\label{sec:numerical}

In this section, we illustrate the performance of our procedure on synthetic and real-world data sets. The code for all the experiments, described below, is available at \href{https://github.com/ArturGoldman/FALCON_changepoint_detection}{Github}\footnote{https://github.com/ArturGoldman/FALCON\_changepoint\_detection}. We consider two choices of the online convex optimization algorithm $\ca$ in Algorithm \ref{alg:fast_change_point}: FTAL and ONS. Recall that we use a class of linear functions of form $\{\theta^\top \psi(x):\theta\in\Theta\}$. We are free to choose a function $\psi(x)$ appropriate for each specific experiment. Hence, $\psi(x)$ may be interpreted as a choice of a feature design for the given data.
We use linear combinations of Hermite or Fourier polynomials of degree $p$ with vectors of coefficients lying in the Euclidean ball $\Theta = \mathcal B(0, 10)$. Finally, for a fair comparison and stability of Algorithm \ref{alg:fast_change_point} across experiments we enforce a property of data both in train and test sets to satisfy $\| \psi(X)\| \leq R$. Specifically, we set $R=1$ by normalizing features in all experiments.

The performance of our method is compared with two popular nonparametric change point detection methods: KLIEP \citep{sugiyama08, liu13} and kernel change point detection with M-statistic \citep{li15}. We also added the comparison with CUSUM version as defined in \cite[Definition 1]{wang20}
in the experiment with a shift in expectation. KLIEP is a density-ratio-based change point detection method, estimating of the KL-divergence between the pre-change and post-change distributions. As we discussed in Section \ref{sec:alg_description}, our approach is related to density-ratio-based methods, so it is reasonable to compare our algorithm with one of them. In \cite{li15}, the authors use kernel methods to approximate the squared maximum mean discrepancy (MMD) between the pre-change and post change distributions. We use a different divergence measure, based on the maximum cross-entropy, but the core idea of maximizing discrepancy between pre-change and post change observations is quite similar. Both KLIEP and M-statistic require a bandwidth parameter $b$ for their computation. In our experiments, we tune this parameter in a way to minimize the detection delay, provided that the number of false alarms or the running length is acceptable.

\subsection{Synthetic data sets}
\label{sec:synth_exp}

The experiments with synthetic data check the ability of the procedure to detect changes in mean and variance in Gaussian sequences. We considered the following setups.

\smallskip

\noindent{\it Example 1: mean shift detection in a Gaussian sequence model.}\quad
We generated a univariate Gaussian sequence of length $150$. The first $75$ observations had the Gaussian distribution $\cn(0, \sigma^2)$ with $\sigma = 0.1$ and the other $75$ were i.i.d. $\cn(\mu, \sigma^2)$ with $\mu = 0.2$ and the same $\sigma$.

\smallskip

\noindent{\it Example 2: variance change detection in a Gaussian sequence model.}\quad
In the second example, we sampled $75$ independent Gaussian random variables $\cn(0, \sigma_0^2)$ with $\sigma_0 = 0.1$ and $75$ random variables with the distribution $\cn(0, \sigma^2)$, $\sigma = 0.3$, so the expectation of all the random variables was the same. CUSUM is not applicable in this case, because it is designed to detect a mean shift.

\begin{table}[ht]
\caption{The thresholds $\z$ and the values of hyperparameters of the competing algorithms in the experiments on synthetic data sets.}
\label{tabl:synthetic_data_thresholds}
\noindent\centering
\resizebox{0.85\linewidth}{!}{
\begin{tabular}{ccccc}
 {\bf METHOD} & \multicolumn{2}{c}{\bf EXAMPLE 1} & \multicolumn{2}{c}{\bf EXAMPLE 2} \\
 & $\z$ & {\bf PARAMETER} & $\z$ & {\bf PARAMETER}  \\
 \hline
Algorithm \ref{alg:fast_change_point} + ONS & $0.3557$ & \makecell{$p = 1, \beta = 0.1, \varepsilon=0.1$, \\ design = Hermite} & $0.7752$ & \makecell{$p = 2, \beta = 0.01, \varepsilon=0.01$, \\ design = Fourier} \\
Algorithm \ref{alg:fast_change_point} + FTAL & $1.729$ & \makecell{$p = 1, \beta = 5$, \\ design = Hermite} & $0.6394$ & \makecell{$p = 2, \beta = 100$, \\ design = Fourier} \\
KLIEP & $6.03$ & $b = 0.2$ & $4.16$ & $b = 0.33$ \\
M-statistic & $9.59$ & $b = 0.5$ & $36.75$ & $b = 0.1$ \\
CUSUM & $0.45$ & - & - & - \\
\end{tabular}
}
\label{tabl:3}
\end{table}

Before we move to the description of our results, let us first elaborate on the threshold tuning procedure.
We sample $T = 150$ i.i.d. samples according to $\sfp$ and compute the maximal value of the corresponding test statistic $\smash{\widehat\cs_t^{(1)}}$, $1 \leq t \leq 150$. For the number of repetitions $J$ we repeat the procedure several times and obtain the values $\max_{1 \leq t \leq T} \smash{\widehat{\cs}_t^{(2)}}, \dots, \max_{1 \leq t \leq T} \smash{\widehat\cs_t^{(J)}}$. Then we put 
\[
    \z = \max\limits_{1 \leq j \leq J} \max\limits_{1 \leq t \leq T} \widehat\cs_t^{(j)}.
\]
Such a choice ensures that the running length of our procedure is not smaller than $T = 150$ with probability at least $1 - 1/(J + 1)$. Indeed, if we run the procedure in the stationary regime and compute the corresponding values of the test statistic $\smash{\widehat\cs_t}$, then the probability that $\max_{1 \leq t \leq T} \smash{\widehat\cs_t}$ exceeds $\z = \max_{1 \leq j \leq J} \max_{1 \leq t \leq T} \smash{\widehat\cs_t^{(j)}}$ is the same as $\max_{1 \leq t \leq T} \smash{\widehat\cs_t^{(j)}}$ exceeds
\[
    \max\left\{\max\limits_{1 \leq t \leq T} \widehat\cs_t, \max\limits_{k \neq j} \max\limits_{1 \leq t \leq T} \widehat\cs_t^{(k)} \right\}.
\]
Since all such probabilities sum to one, we conclude that $\p( \max_{1 \leq t \leq T} \widehat\cs_t > \z) = 1 / (J + 1)$,
provided that there are no change points.
We took $J = 9$ in the experiments with changes in mean and in variance.
The information about thresholds in the experiments with artificial data is collected in Table \ref{tabl:synthetic_data_thresholds}.

\begin{figure}[ht]
    \centering
    \subfloat{
        \includegraphics[width=0.48\textwidth]{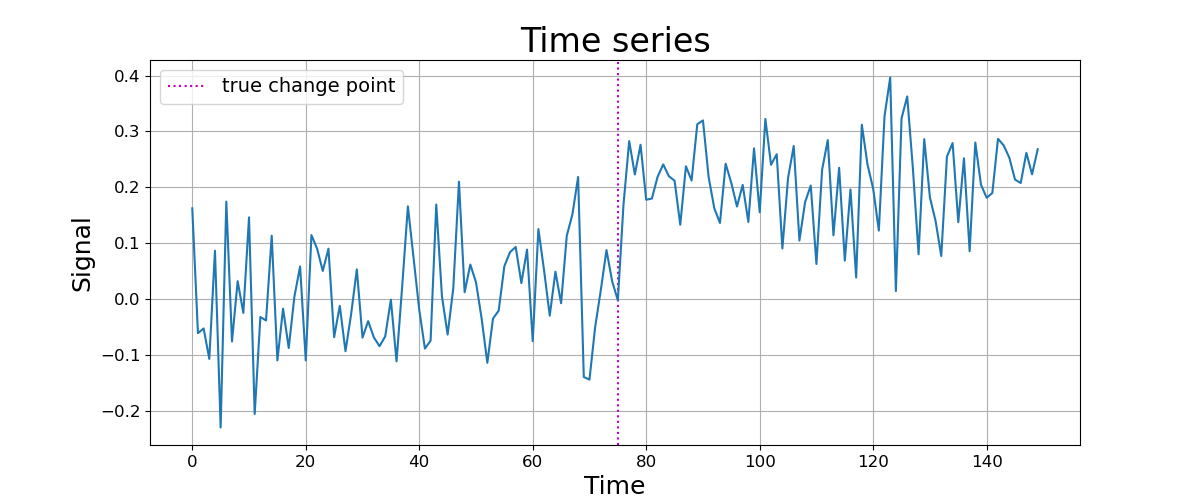}
    }
    \hfill
    \subfloat{
        \includegraphics[width=0.48\textwidth]{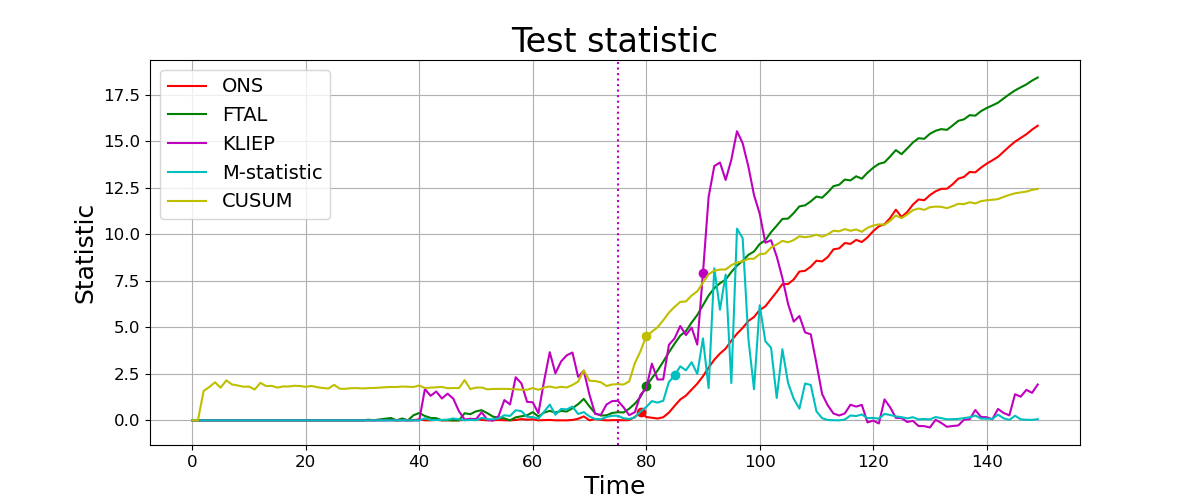}
    }
    \caption{An example of change point detection on synthetic data set for a mean shift detection in a Gaussian sequence model. Left: the sequence of observations. Right: corresponding values of test statistics $\smash{\widehat\cs_t}$ for Algorithm \ref{alg:fast_change_point} with two variants of the algorithm $\ca$ (ONS (red) and FTAL (green)), CUSUM for mean shift detection (yellow), KLIEP (magenta) and M-statistics (cyan). The dashed vertical line corresponds to the true change point $\tstar$. The circle markers on solid lines correspond to the detection moments.}
    \label{fig:synthetic}
\end{figure}

The setup was as follows.
In each example, we sampled an artificial sequence $10$ times and computed the average detection delays for Algorithm \ref{alg:fast_change_point} with choice of $\ca$ and feature design for $\psi$ as specified in Table \ref{tabl:3}, and for the competitors (CUSUM, KLIEP and kernel change point detection with M-statistic) for each realization. The results are displayed in Table \ref{tabl:1} and Figure \ref{fig:synthetic}. During the first $30$ iterations, we collected the observations for further training while also gathering the statistics for the components of OCO algorithm $\ca$ corresponding to $\smash{\widehat\theta_{\tau,t}}=0$ (which corresponds to the optimal value in case no changepoint has occurred), but the test statistic $\smash{\widehat\theta_{\tau,t}}$ itself was not computed. We also slightly adjusted the test statistic $\smash{\widehat\cs_t}$: instead of maximizing $\smash{\widehat \ct_{\tau, t}}$ over the whole set $\{1, \dots, t-1\}$, we took the maximum with respect to $\tau \in \{10, 11, \dots, t - 10\}$.
This simple trick helped us reduce the detection delay. The hyperparameters of KLIEP and M-statistic-based kernel change point methods were tuned in a way to minimize the average detection delay while keeping the running length at least $150$ with high probability.

\begin{table}[ht]
\caption{Detection delays of Algorithm \ref{alg:fast_change_point} with two variants of the algorithm $\ca$ (ONS and FTAL), KLIEP, kernel change point with M-statistic, and CUSUM on synthetic data sets. Two best results are boldfaced.}
\noindent\centering
\begin{tabular}{lll}
{\bf METHOD} & {\bf EXAMPLE 1} & {\bf EXAMPLE 2}  \\
\hline
Algorithm \ref{alg:fast_change_point} + ONS & $6.9 \pm 3.9$ & $\mathbf{11.2 \pm 6.1}$ \\
Algorithm \ref{alg:fast_change_point} + FTAL & $\mathbf{5.9 \pm 1.9}$ & $\mathbf{15.9 \pm 9.3}$ \\
KLIEP & $8.9 \pm 3.6$ & $19.2 \pm 18.4$\\
M-statistic & $10.4 \pm 3.4$ & $51.1 \pm 27.3$ \\
CUSUM & $\mathbf{5.0 \pm 2.0}$ & -- \\
\end{tabular}
\label{tabl:1}
\end{table}
According to Table \ref{tabl:1}, Algorithm \ref{alg:fast_change_point} is the most efficient method to detect a change point amongst competitors. It only loses to CUSUM in the mean shift detection example. That is not surprising, because CUSUM was especially designed to detect changes in mean of a Gaussian sequence. We move to the experiments on the real-world data sets.

\subsection{Univariate data: speech records analysis}

\begin{figure}[ht]
    \noindent\centering
    \includegraphics[width=0.5\linewidth]{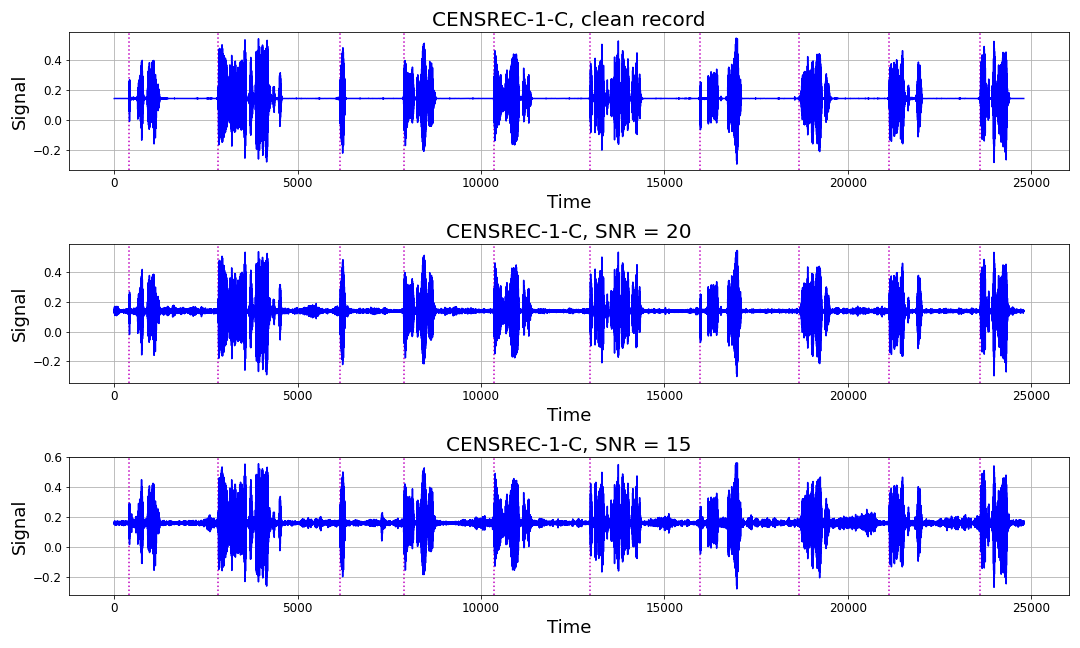}
    \caption{The CENSREC-1-C speech records with different amount of noise.}
    \label{fig:CENSREC}
\end{figure}

We used CENSREC-1-C\footnote{http://research.nii.ac.jp/src/en/CENSREC-1-C.html} data in the Speech Resource Consortium (SRC) corpora provided by National Institute of Informatics (NII) to test the algorithm in practical tasks.
The data set contains a clean speech record (MAH\_clean) and the same record corrupted with noise of different magnitude (MAH\_N1\_SNR20, MAH\_N1\_SNR15). The signal plots are showed in Figure \ref{fig:CENSREC}. We preprocessed the data as follows. First, we normalized the data. Next, we chose $10$ segments with a single change from silence/noise to speech, and then each $10$-th observation was taken. The first four segments were used to tune the hyperparameters and the thresholds and the other six were for testing. The true change point values were set on the MAH\_clean data set and used in the noisy versions of the record. 

In Algorithm \ref{alg:fast_change_point}, we used linear combinations of Fourier or Hermite polynomials of degree less than $3$. We also demonstrate in this experiment how one can choose best performing data features as a part of hyperparameter tuning over validation set. The bandwidths used in KLIEP and the M-statistic based algorithm are shown in Table \ref{tabl:4}. It also contains the corresponding values of thresholds.
We computed detection delays for each algorithm on each of $6$ test segments. The results are reported in Table \ref{tabl:2}.

\begin{table}[ht]
\caption{The thresholds $\z$ and the values of hyperparameters of the competing algorithms in the experiments on the CENSREC-1-C data set.}
\noindent\centering
\resizebox{\linewidth}{!}{
\begin{tabular}{ccccccc}
 & \multicolumn{2}{c}{\bf CLEAN RECORD} & \multicolumn{2}{c}{\bf SNR20} & \multicolumn{2}{c}{\bf SNR15} \\
 & $\z$ & {\bf PARAMETER} & $\z$ & {\bf PARAMETER} & $\z$ & {\bf PARAMETER} \\
 \hline
Algorithm \ref{alg:fast_change_point} + ONS & $0.0003$ & \makecell{$p = 2, \beta = 0.56, \varepsilon=0.01$,\\ design = Fourier} & $0.0001$ & \makecell{$p = 2, \beta = 0.56, \varepsilon=1$,\\ design = Hermite} & $0.005$ & \makecell{$p = 2, \beta = 0.56, \varepsilon=1$,\\ design = Fourier} \\
Algorithm \ref{alg:fast_change_point} + FTAL & $0.048$ & \makecell{$p = 2, \beta = 10$,\\ design = Fourier} & $0.01$ & \makecell{$p = 2, \beta = 20$,\\ design = Hermite} & $0.05$ & \makecell{$p = 2, \beta = 2.5$,\\ design = Fourier} \\
KLIEP & $0.61$ & $b = 0.2$ & $1.17$ & $b = 0.075$ & $0.079$ & $b = 0.1$ \\
M-statistic & $4.68 \cdot 10^{-3}$ & $b = 0.1$ & $15.7 \cdot 10^{-3}$ & $b = 0.5$ & $10^{-4}$ & $b = 2$ \\
\end{tabular}
}
\label{tabl:4}
\end{table}

\begin{table}[ht]
\caption{Average detection delays (DD) and the number of false alarms (FA) of Algorithm \ref{alg:fast_change_point} (with choices of ONS and FTAL for $\ca$), KLIEP, and kernel change point detector with M-statistic on the CENSREC-1-C speech records with different noise level. Two best results are boldfaced.}

\hspace{-0.5in}
\noindent\centering
\begin{tabular}{ccccccc}
 & \multicolumn{2}{c}{\bf CLEAN RECORD} & \multicolumn{2}{c}{\bf SNR20} & \multicolumn{2}{c}{\bf SNR15} \\
 & {\bf FA} & {\bf DD} & {\bf FA} & {\bf DD} & {\bf FA} & {\bf DD} \\
 \hline
Algorithm \ref{alg:fast_change_point} + ONS
& $0$ & $\mathbf{6.7 \pm 3.8}$ & $0$ & $\mathbf{13.3 \pm 18.0}$ & $0$ & $\mathbf{14.0 \pm 15.7}$ \\
Algorithm \ref{alg:fast_change_point} + FTAL
& $0$ & $9.5 \pm 17.4$ & $0$ & $\mathbf{14.8\pm 17.2}$ & $1$ & $\mathbf{10.2 \pm 10.0}$ \\
KLIEP & $0$ & $10.3 \pm 19.2$ & $0$ & $21.0 \pm 21.2$ & $0$ & $20.5 \pm 21.6$ \\
M-statistic & $0$ & $\mathbf{7.3 \pm 13.1}$ & $0$ & $17.3 \pm 20.1$ & $0$ & $14.8 \pm 19.4$ \\
\end{tabular}
\label{tabl:2}
\end{table}
It can be seen that Algorithm \ref{alg:fast_change_point} manages to show the best and sometimes second best performance for various choices of $\ca$. In contrast to the experiments on synthetic data sets, KLIEP behaves poorly in this example.

\subsection{Multivariate data I: activity change recognition}
\label{sec:activity_change_recognition}

\begin{figure}[ht]
    \centering
    \begin{minipage}{0.55\linewidth}
        \noindent
        \center{
        \includegraphics[width=\linewidth]{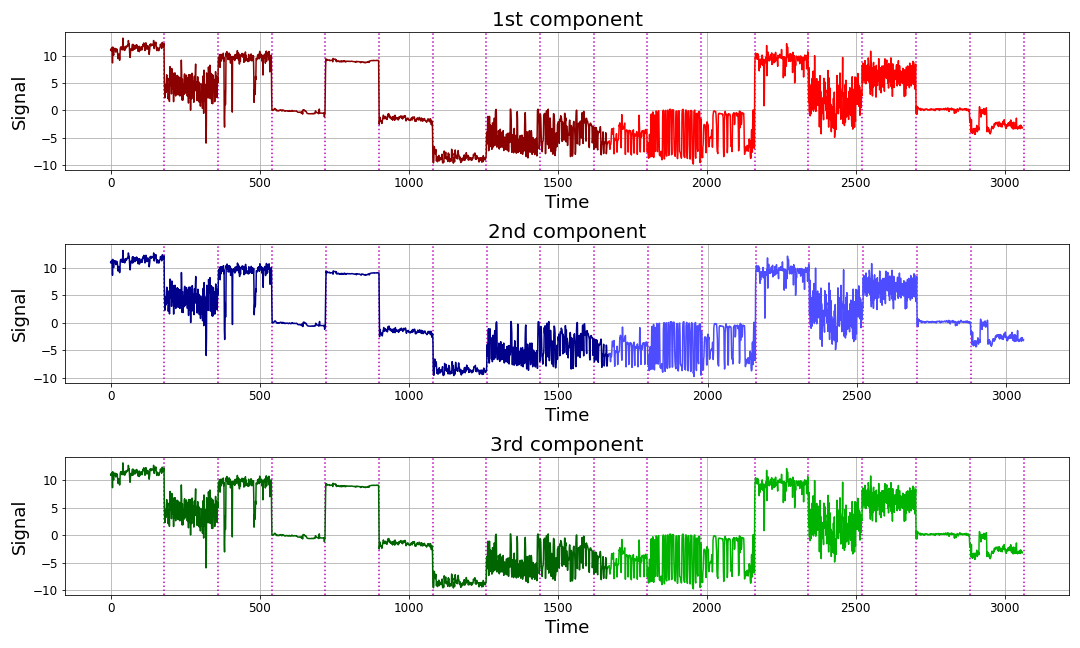}
        }
    \end{minipage}
    \hspace{1cm}
    \begin{minipage}{0.33\linewidth}
        \noindent
        \center{
        \includegraphics[width=\linewidth]{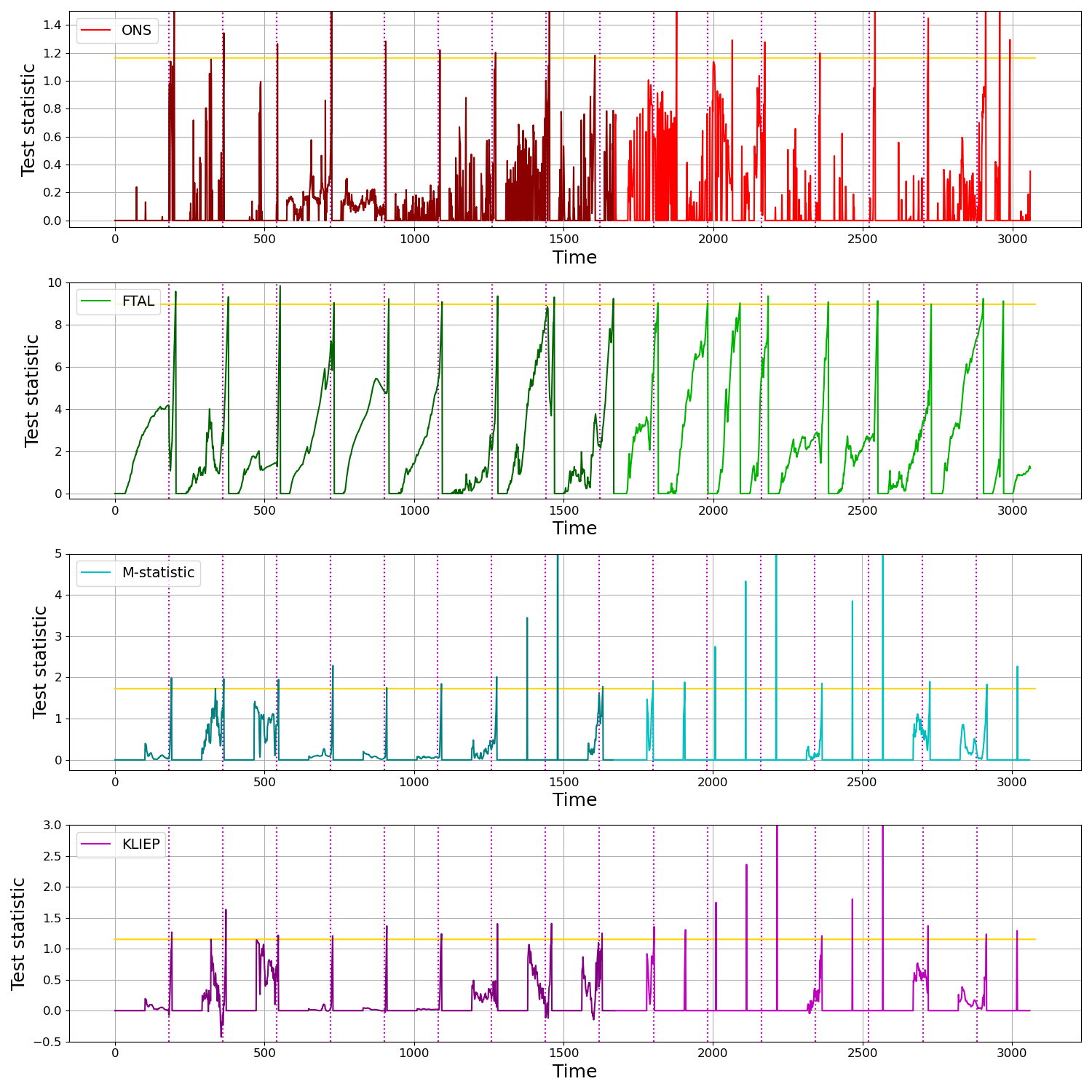}
        }
    \end{minipage}
    \caption{Left: three-dimensional time series from the WISDM data set. Right: the corresponding values of the test statistics for Algorithm \ref{alg:fast_change_point} (with two variants of the algorithm $\ca$, red and green), the kernel change point detector with M-statistic (cyan) and KLIEP (magenta). The dotted vertical lines and the solid yellow ones correspond to the change points and thresholds, respectively. The dark shades stand for the training period, while the light ones denote the test part.}
    \label{fig:WISDM}
\end{figure}

In this section, we apply Algorithm \ref{alg:fast_change_point} to detect changes in a user's physical activity. In our experiments, we took a part of the data set WISDM \citep{weiss19}, containing 3-dimensional measurements of a smartphone accelerometer, measured at a rate 20Hz. We preprocessed the data set, taking only each 20-th observation. Nevertheless, even after such a reduction the length of the time series was over $3000$. The observations are displayed in Figure \ref{fig:WISDM}. During the measurement period, the user changed a kind of activity $17$ times, i.e. the time series contained $17$ change points. Our goal was to detect them as soon as possible.

We applied Algorithm \ref{alg:fast_change_point} with a feature vector $\smash{\psi(x) = (1, x^\top)^\top}$, which essentially returns the initial multidimensional vector with an added bias term for both ONS and FTAL choices for $\ca$.
As before, we compared our procedures with KLIEP and the kernel change point detector with M-statistic. The information about thresholds and parameters of the algorithms is collected in Table \ref{tabl:wisdm}. We split the data set into train and test parts in such a way that the former one contains $8$ change points. We set the thresholds as the maximal value of the test statistics on the first four stationary parts of the time series. After that, we computed the average detection delay of each algorithm.
The results are presented in Table \ref{tabl:wisdm}. The plots of the test statistics are shown in Figure \ref{fig:WISDM}. According to Table \ref{tabl:wisdm}, Algorithm \ref{alg:fast_change_point} with choice of algorithm $\ca$ (ONS or FTAL) outperforms competitors, having a shorter detection delay and a smaller number of false alarms.

\begin{table}[ht]
\caption{The number of false alarms (FA) and the average detection delays (DD) of Algorithm \ref{alg:fast_change_point} (with two variants of the algorithm $\ca$), KLIEP, and the kernel change point detector with M-statistic on the WISDM data set. Two best results are boldfaced.}
\vspace{0.2in}
\noindent\centering
\begin{tabular}{lllll}
{\bf METHOD} & $\z$ & {\bf PARAMETER} & {\bf FA} & {\bf DD} \\
\hline
Algorithm \ref{alg:fast_change_point} + ONS & $1.16$ & $p = 1, \beta = 0.01, \varepsilon=0.01$ & $2$ & $\mathbf{24.3 \pm 29.5}$\\
Algorithm \ref{alg:fast_change_point} + FTAL & $8.96$ & $p = 1, \beta = 10$ & $1$ & $\mathbf{25.4 \pm 30.7}$\\
KLIEP & $1.15$ & $b = 20$ & 4 & $30.9 \pm 30.1$\\ 
M-statistic & $1.73$ & $b = 20$ & 4 & $30.7 \pm 28.2$\\
\end{tabular}
\label{tabl:wisdm}
\end{table}

\subsection{Multivariate data II: room occupancy detection}
\label{subsec:dataset_occupancy}

\begin{figure}[H]
    \centering
    \subfloat{
        \includegraphics[width=0.48\textwidth]{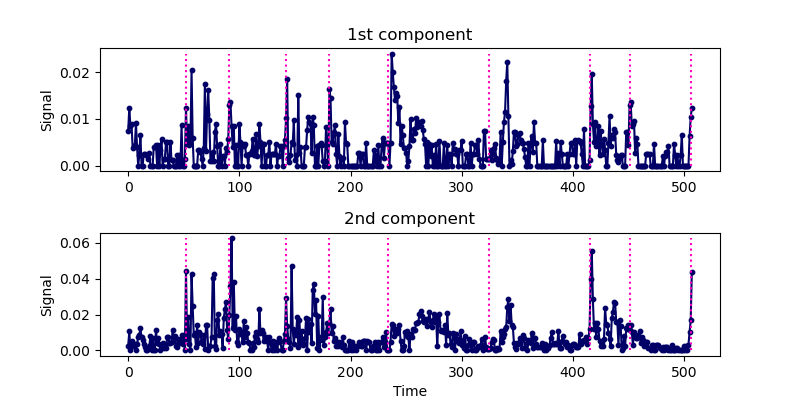}
    }
    \hfill
    \subfloat{
        \includegraphics[width=0.48\textwidth]{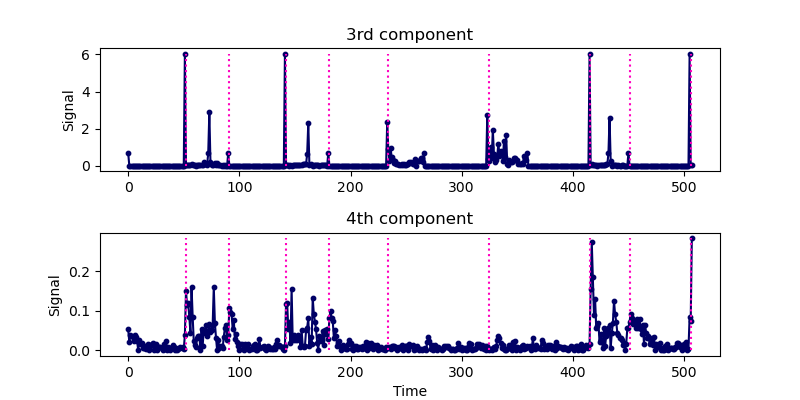}
    }
    \caption{The four-dimensional time series from the Occupancy data set.}
    \label{fig:occupancy_data}
\end{figure}

Finally, we applied Algorithm \ref{alg:fast_change_point} to detect changes in room occupancy based on Temperature, Humidity, Light, and ${\rm CO}_2$ variables. The four-dimensional time series was obtained from UCI repository. The data preprocessing pipeline comprised two sequential steps. Firstly, we selected every 16th observation to reduce the length of the time series. Additionally, we calculated the differences between the logarithms of consecutive observations and normalized these differences by dividing them by the earliest observation. Last transformation aimed to convert the non-stationary time series into a stationary one. After the preprocessing, the time series consisted of roughly 500 data points, with 9 of them labeled as change points. Description of the change point annotation pipeline may be found in \citep[Section Annotation Collection]{burg2022evaluation}. The time series is displayed in Figure \ref{fig:occupancy_data}.

For a choice of $\psi(x)$ in Algorithm \ref{alg:fast_change_point} we took the same function as in the previous case: $\smash{\psi(x)=(1,x^\top)^\top}$.
We compared the performance of Algorithm \ref{alg:fast_change_point} with ONS and FTAL for the choice of $\ca$ with KLIEP and the kernel change point detector with M-statistic. 
A reader can find bandwidths for KLIEP, M-statistic and thresholds for all methods in Table \ref{tabl:occupancy_res}, as well as the number of false alarms and detection delays on the test part. 
Algorithm \ref{alg:fast_change_point} demonstrates the shortest delay with a small number of false alarms for either choice of ONS or FTAL as an online optimization algorithm $\ca$.

\begin{table}[H]
\caption{The number of false alarms (FA) and the average detection delays (DD) of Algorithm \ref{alg:fast_change_point} (with two variants of the algorithm $\ca$), KLIEP, and the kernel change point detector with M-statistic on the occupancy data set. Two best results are boldfaced.}
\noindent\centering
\begin{tabular}{lllll}
{\bf METHOD} & {$\z$} & {\bf PARAMETER} & {\bf FA} & {\bf DD} \\
\hline
Algorithm \ref{alg:fast_change_point} + ONS & 0.09 & p = 1,\ $\beta$ = 0.05, $\varepsilon=1$ & 1 & $\mathbf{6.0 \pm 2.8}$ \\
Algorithm \ref{alg:fast_change_point} + FTAL & 0.08 & p = 1,\ $\beta$ = 1 & 2 & $\mathbf{2.5 \pm 2.2}$ \\
M-statistic & 4 & b = 0.2 & $1$ & $10.25 \pm 5.67$\\
KLIEP & 1.66 & b = 0.5 & $1$ & $11.25 \pm 6.68$\\ 
\end{tabular}
\label{tabl:occupancy_res}
\end{table}

\bibliographystyle{abbrvnat}
\bibliography{references.bib}

\appendix

\section{Proofs of the results from Section \ref{sec:alg_description}}
\label{sec:lemmata_proofs}

This section collects proofs of the results, presented in Section \ref{sec:alg_description}.

\subsection{Proof of Lemma \ref{lem:js}}

Let
\[
    f^*(x) = \log\frac{\sfp(x)}{\sfq(x)},
    \quad
    D^*(x) = \frac{e^{f^*(x)}}{1 + e^{f^*(x)}} = \frac{\sfp(x)}{\sfp(x) + \sfq(x)},
\]
and
\[
    \ct^* = \frac{\tstar (t - \tstar)}t
    \Bigg[ \frac1{\tstar} \sum\limits_{s=1}^\tau \log\big( 2D^*(X_s) \big)
    + \frac1{t - \tstar} \sum\limits_{s = \tau + 1}^t \log\big( 2 - 2D^*(X_s) \big) \Bigg].
\]
It is straightforward to check that the expectation of $\ct^*$ is proportional to the Jensen-Shannon divergence between $\sfp$ and $\sfq$. To be more specific, it holds that
\begin{align*}
    \E \ct^*
    &
    = \frac{\tstar(t - \tstar)}t \left[ \int \log(2 D^*(x)) \sfp(x) \dd \m + \int \log(2 - 2 D^*(x)) \sfq(x) \dd \m \right]
    \\&
    = \frac{\tstar(t - \tstar)}t \left[ \int \log\left( \frac{2 \sfp(x)}{\sfp(x) + \sfq(x)} \right) \sfp(x) \dd \m + \int \log\left( \frac{2 \sfq(x)}{\sfp(x) + \sfq(x)} \right) \sfq(x) \dd \m \right]
    \\&
    = \frac{2\tstar(t - \tstar)}t \left[ \KL\left(\sfp, \frac{\sfp + \sfq}2 \right) + \KL\left(\sfq, \frac{\sfp + \sfq}2 \right) \right]
    \equiv \frac{2\tstar(t - \tstar) \JS(\sfp, \sfq)}t.
\end{align*}
On the other hand, introducing $D_\theta(x) = e^{\theta^\top \psi(x)} / (1 + e^{\theta^\top \psi(x)})$, we observe that
\begin{align*}
    \E \ct^* - \E \ct_{\tstar, t}(\theta)
    &
    = \frac{\tstar(t - \tstar)}t \left[ \int \log\left( \frac{D^*(x)}{D_\theta(x)}\right) \sfp(x) \dd \m + \int \log\left( \frac{1 - D^*(x)}{1 - D_\theta(x)}\right) \sfq(x) \dd \m \right]
    \\&
    = \frac{\tstar(t - \tstar)}t \int \log\left(\frac{D^*(x) / \big(1 - D^*(x)\big)}{D_\theta(x) / \big(1 - D_\theta(x)\big)}\right) \sfp(x) \dd \m
    \\&\quad
    + \frac{\tstar(t - \tstar)}t \int \log\left( \frac{1 - D^*(x)}{1 - D_\theta(x)}\right) \big(\sfp(x) + \sfq(x) \big) \dd \m.
\end{align*}
Since, by the definition, $\sfp(x) = D^*(x) \big(\sfp(x) + \sfq(x) \big)$ and $D^*(x) = \big(1 - D^*(x)\big) e^{f^*(x)}$,
we obtain that 
\begin{align}
    \label{eq:fstar_f_diff}
    \E \ct^* - \E \ct_{\tstar, t}(\theta)
    &\notag
    = \frac{\tstar(t - \tstar)}t \left[ \int \frac{e^{f^*(x)}}{1 + e^{f^*(x)}} \big(f^*(x) - \theta^\top \psi(x) \big) \big(\sfp(x) + \sfq(x) \big) \dd \m \right]
    \\&\quad
    - \frac{\tstar(t - \tstar)}t \left[ \int \log\left(\frac{1 + e^{f^*(x)}}{1 + e^{\theta^\top \psi(x)}}\right) \big(\sfp(x) + \sfq(x) \big) \dd \m \right].
\end{align}
Consider a function $g : \R^2 \rightarrow \R$, defined as
\[
    g(u, v) = \frac{(u - v) e^u}{1 + e^u} - \log\left( \frac{1 + e^u}{1 + e^v} \right).
\]
Note that, for any $u, v \in \R$, we have $g(u, u) = 0$,
\[
    \left.\frac{\partial g(u, v)}{\partial v} \right|_{v = u} = \left.\left[- \frac{e^u}{1 + e^u} + \frac{e^v}{1 + e^v}\right]\right|_{v=u} = 0,
    \quad \text{and} \quad
    \frac{\partial^2 g(u, v)}{\partial v^2}
    = \frac{e^v}{(1 + e^v)^2}
    \leq \frac14.
\]
Hence, for any $u, v\in \R$, it holds that
\[
    g(u, v) \leq \frac{(u - v)^2}8.
\]
Applying this inequality to the right-hand side of \eqref{eq:fstar_f_diff}, we obtain that
\begin{align*}
    \E \ct^* - \E \ct_{\tstar, t}(\theta)
    &
    \leq \frac{\tstar (t - \tstar)}{8t} \left[ \int \big(\theta^\top \psi(x) - f^*(x)\big)^2 \big(\sfp(x) + \sfq(x) \big) \dd \m \right]
    \\&
    = \frac{\tstar (t - \tstar)}{4t} \left\|\theta^\top \psi - \log(\sfp / \sfq) \right\|_{L_2((\sfp + \sfq) / 2)}^2.
\end{align*}
Taking into account that $\E \ct^* = 2\tstar(t - \tstar) \JS(\sfp, \sfq) / t$, we finally get
\[
    \E \ct_{\tstar, t}(\theta) \geq \frac{2\tstar(t - \tstar)}t \left( \JS(\sfp, \sfq) - \frac1{8} \left\|\theta^\top \psi - \log(\sfp / \sfq) \right\|_{L_2((\sfp + \sfq) / 2)}^2 \right).
\]
\myendproof

\subsection{Proof of Lemma \ref{lem:exp-concave}}
\label{sec:lem_exp-concave_proof}

It is enough to check that $\nabla^2 \varphi_{\tau, t}(\theta) \succeq 0.5 e^{-B} \nabla \varphi_{\tau, t}(\theta) \nabla \varphi_{\tau, t}(\theta)^\top$ for any $\theta \in \Theta$, that is,
\[
    \left( v^\top \nabla \varphi_{\tau, t}(\theta) \right)^2 \leq 2 e^{B} \; v^\top \nabla^2 \varphi_{\tau, t}(\theta) v
    \quad \text{for any $v \in \R^d$ and any $\theta \in \Theta$.}
\]
Let us fix an arbitrary $\theta \in \Theta$ and introduce
\[
    \alpha_s =
    \begin{cases}
        \tau^{-1} \cdot e^{-\theta^\top \psi(X_s)} \slash (1 + e^{-\theta^\top \psi(X_s)}),
        \quad \text{if $1 \leq s \leq \tau$,}\\
        e^{\theta^\top \psi(X_t)} \slash (1 + e^{\theta^\top \psi(X_t)})
        \quad \text{if $s = t$.}
    \end{cases}
\]
Due to the definition of $\varphi_{\tau, t}(\theta)$ (see \eqref{eq:phi}), we have
\[
    \nabla \varphi_{\tau, t}(\theta)
    = - \frac1{\tau} \sum\limits_{s = 1}^\tau \frac{e^{-\theta^\top \psi(X_s)}}{1 + e^{-\theta^\top \psi(X_s)}} \, \psi(X_s)
    + \frac{e^{\theta^\top \psi(X_t)}}{1 + e^{\theta^\top \psi(X_t)}} \, \psi(X_t)
    = \alpha_t \psi(X_t) - \sum\limits_{s = 1}^\tau \alpha_s \psi(X_s)
\]
and
\begin{align*}
    \nabla^2 \varphi_{\tau, t}(\theta)
    &
    = \frac1{\tau} \sum\limits_{s = 1}^\tau \frac{e^{-\theta^\top \psi(X_s)}}{(1 + e^{-\theta^\top \psi(X_s)})^2} \, \psi(X_s) \psi(X_s)^\top
    + \frac{e^{\theta^\top \psi(X_t)}}{(1 + e^{\theta^\top \psi(X_t)})^2} \, \psi(X_t)\psi(X_t)^\top
    \\&
    \succeq \frac1{1 + e^{B}} \left( \alpha_t \psi(X_t)\psi(X_t)^\top + \sum\limits_{s = 1}^\tau \alpha_s \psi(X_s) \psi(X_s)^\top \right).
\end{align*}
Using the Cauchy-Schwarz inequality, we obtain that
\begin{align*}
    \left( v^\top \nabla \varphi_{\tau, t}(\theta) \right)^2
    &
    = \left( \alpha_t v^\top \psi(X_t) - \sum\limits_{s = 1}^\tau \alpha_s v^\top \psi(X_s) \right)^2
    \\&
    \leq \left(\alpha_t + \sum\limits_{s = 1}^\tau \alpha_s \right) \left( \alpha_t \left(v^\top \psi(X_t) \right)^2 + \sum\limits_{s = 1}^\tau \alpha_s \left(v^\top \psi(X_s)\right)^2 \right) 
    \\&
    \leq \left(\frac{e^{B}}{1 + e^{B}} + \sum\limits_{s = 1}^\tau \frac{e^{B}}{\tau (1 + e^{B})} \right) \left( \alpha_t \left(v^\top \psi(X_t) \right)^2 + \sum\limits_{s = 1}^\tau \alpha_s \left(v^\top \psi(X_s)\right)^2 \right) 
    \\&
    \leq 2 e^{B} \; v^\top \nabla^2 \varphi_{\tau, t}(\theta) v
\end{align*}
for any $v \in \R^d$. The proof is finished.

\myendproof

\section{Proof of Theorem \ref{th:ftal_test_stat_upper_bound}}
\label{sec:th_ftal_test_stat_upper_bound_proof}

Introducing
\begin{align}
    \label{eq:scp_def}
    \scp_\tau(\theta) &= -\frac{1}{\tau}\sum_{s = 1}^\tau \left( \log\left(\frac{1 + e^{-\theta^\top \psi(X_s)}}{2}\right) + \frac{1}{2}\theta^\top \psi(X_s) \right),
    \\
    \label{eq:scq_def}
    \scq_{\tau, t} &= \sum_{s = \tau + 1}^t\left( -\log\left(\frac{1 + e^{\widehat{\theta}_{\tau, s - 1}^\top \psi(X_s)}}{2}\right) + \frac{1}{2}\widehat{\theta}_{\tau, s - 1}^\top\psi(X_s) \right), 
\end{align}
we observe that
\begin{align*}
    \widehat\ct_{\tau, t}
    &
    = - \frac{\tau}{t} \sum\limits_{s = 1}^t \varphi_{\tau, s}(\widehat\theta_{\tau, s - 1})
    \\&
    = -\frac{\tau}{t} \sum\limits_{s = \tau + 1}^t \frac1\tau \sum\limits_{m = 1}^\tau \log \frac{1 + \exp\{-\widehat\theta_{\tau, s - 1}^\top \psi(X_m)\}}2
    - \frac{\tau}t \sum\limits_{s = \tau + 1}^t \log \frac{1 + \exp\{\widehat\theta_{\tau, s - 1}^\top \psi(X_s)\}}2
    \\&
    = \frac{\tau}t \sum\limits_{s = \tau + 1}^t \scp_\tau(\widehat\theta_{\tau, s - 1}) + \frac{\tau}t \scq_{\tau, t}.
\end{align*}
For any $\theta \in \Theta$ let us denote
\begin{align}
\label{eq:scp_mean_def}
    \overline\scp_\tau(\theta)
    = \E_{X' \sim \sfp}\left[-\log\left(\frac{1 + e^{-\theta^\top \psi(X')}}{2}\right) - \frac{1}{2}\theta^\top \psi(X')\right] ,
\end{align}
and define
\begin{align}
\label{eq:scq_mean_def}
    \overline\scq_{\tau, t}
    = \sum_{s = \tau + 1}^t \E_{X' \sim \sfp}\left[ -\log\left(\frac{1 + e^{\widehat{\theta}_{\tau, s - 1}^\top \psi(X')}}{2}\right) + \frac{1}{2}\widehat{\theta}_{\tau, s - 1}^\top\psi(X') \right],
\end{align}
where $X'$ is independent of $X_1, \dots, X_t$.
It is evident that
\begin{align}
\label{eq:tilde_T_decomp}
    \frac{t}{\tau}\widehat\ct_{\tau, t}
    &\notag
    = \frac12 \sum\limits_{s = \tau + 1}^t \overline\scp_\tau(\widehat\theta_{\tau, s - 1})
    + \overline\scq_{\tau, t}
    \\&\quad
    + \sum\limits_{s = \tau + 1}^t \left( \scp_\tau(\widehat\theta_{\tau, s - 1}) - \frac12 \overline\scp_\tau(\widehat\theta_{\tau, s - 1})\right)
    + \left( \scq_{\tau, t} - \overline\scq_{\tau, t} \right).
\end{align}
Let
\begin{align}
\label{eq:kappa_Sigma_def}
    \varkappa = \frac{e^B}{(1 + e^B)^2}
    \quad \text{and} \quad
    \Sigma = \E_{X \sim \sfp}\left[\psi(X) \psi(X)^\top \right],
\end{align}
and note that the mapping $u \mapsto -\log(1 + e^{-u}) - u/2$ is $\varkappa$-strongly concave on $[-B, B]$. This, together with the identities $\overline{\scp}_\tau(0) = 0$ and $\nabla \overline{\scp}_\tau(0) = 0$ implies that
\begin{align*}
    \overline{\scp}_\tau(\theta)
    \leq -\frac{\varkappa}{2}\E_{X' \sim \sfp}\left[(\theta^\top \psi(X'))^2\right]
    = -\frac{\varkappa}{2} \left\|\Sigma^{1/2}\theta \right\|^2,
    \quad \text{for all $\theta \in \Theta$} .
\end{align*}
Similarly, we obtain that
\begin{align*}
    \overline{\scq}_{\tau, t} \leq -\frac{\varkappa}{2} \sum_{s = \tau + 1}^t \E_{X' \sim \sfp}\left[(\widehat{\theta}_{\tau, s - 1}^\top \psi(X'))^2\right]
    = -\frac{\varkappa}{2}\sum_{s = \tau + 1}^t \left\|\Sigma^{1/2} \widehat{\theta}_{\tau, s - 1} \right\|^2 .
\end{align*}
This yields that
\[
    \frac12 \sum\limits_{s = \tau + 1}^t \overline\scp_\tau(\widehat\theta_{\tau, s - 1}) + \overline\scq_{\tau, t}
    \leq -\frac{3 \varkappa}{4} \sum\limits_{s = \tau + 1}^t \left\|\Sigma^{1/2} \widehat\theta_{\tau, s - 1} \right\|^2.
\]
Combining this bound with \eqref{eq:tilde_T_decomp}, we conclude that
\begin{align}
\label{eq:tilde_T_tau_t_naive_bound_decomp}
    \notag
    \frac{t}{\tau} \widehat\ct_{\tau, t}
    &
    \leq \sum\limits_{s = \tau + 1}^t \left( \scp_\tau(\widehat\theta_{\tau, s - 1}) - \frac12 \, \overline\scp_\tau(\widehat\theta_{\tau, s - 1})\right)
    \\&\quad
    + \left( \scq_{\tau, t} - \overline\scq_{\tau, t} \right) - \frac{3 \varkappa}{4} \sum\limits_{s = \tau + 1}^t \left\|\Sigma^{1/2} \widehat\theta_{\tau, s - 1} \right\|^2.
\end{align}
The rest of the proof is devoted to analysis of two terms in the right-hand side. We start with the former one. First, we note that
\begin{align}
\label{eq:scp_tau_sum_sup_bound}
    &
    \sum\limits_{s = \tau + 1}^t \left( \scp_\tau(\widehat\theta_{\tau, s - 1}) - \frac12 \, \overline\scp_\tau(\widehat\theta_{\tau, s - 1}) \right)
    \leq (t - \tau) \; \sup\limits_{\theta \in \Theta} \left\{ \scp_\tau(\theta) - \frac12 \, \overline\scp_\tau(\theta) \right\}.
\end{align}
Second, we use local Rademacher complexities to derive a uniform high-probability upper bound on
\[
    \scp_\tau(\theta) - \frac12 \, \overline\scp_\tau(\theta),
    \quad
    \theta \in \Theta.
\]
In particular, using the findings of \cite{bbm05}, we prove the following result.
\begin{Lem}
    \label{lem:scp_uniform_bound}
    Suppose that $|\theta^\top \psi(X)| \leq B$ for all $\theta \in \Theta$ and almost all $X \sim \sfp$.
    Let $\scp_\tau(\theta)$ and $\overline{\scp}_\tau(\theta)$ be as defined in \eqref{eq:scp_def} and \eqref{eq:scp_mean_def}, respectively.
    Then, for any $\delta \in (0, 1)$, with probability at least $(1 - \delta)$ it holds that
    \[
        \scp_\tau(\theta) - \frac12 \, \overline{\scp}_\tau(\theta)
        \leq \frac{3 e^B d}{\tau}
        + \frac{11 B \log(1 / \delta)}{4 \tau}
        + \frac{5 e^B \log(1 / \delta)}{2 \tau}
    \]
    simultaneously for all $\theta \in \Theta$.
\end{Lem}

We postpone the proof of Lemma \ref{lem:scp_uniform_bound} to Appendix \ref{sec:scp_uniform_bound_proof} and move to the study of $\scq_{\tau, t} - \overline\scq_{\tau, t}$.
The analysis of this term is more intricate and relies on the martingale Bernstein inequality \citep{freedman75}.

\begin{Lem}
\label{lem:scq_bound}
    Assume that $|\theta^\top \psi(X)| \leq B$ for all $\theta \in \Theta$ and almost all $X \sim \sfp$.
    Let $\scq_{\tau, t}$ and $\overline{\scq}_{\tau, t}$ be as defined in \eqref{eq:scq_def} and \eqref{eq:scq_mean_def}, respectively.
    Then, for any $\delta \in (0, 1)$, with probability at least $(1 - \delta)$, it holds that
    \[
        \scq_{\tau, t} - \overline{\scq}_{\tau, t} - \frac{3\varkappa}{4} \sum_{s = \tau + 1}^t \|\Sigma^{1/2}\widehat{\theta}_{\tau, s - 1}\|^2 \leq  \left(2B + \frac{8 e^B}3 \right) \log(3 / \delta)
    \]
    simultaneously for all $t \geq \tau + 1$, where $\Sigma$ is given by \eqref{eq:kappa_Sigma_def}.
\end{Lem}
We provide the proof of Lemma \ref{lem:scq_bound} in Appendix \ref{sec:scq_bound_proof} below. Combining Lemma \ref{lem:scp_uniform_bound} and Lemma \ref{lem:scq_bound} with \eqref{eq:tilde_T_tau_t_naive_bound_decomp} and \eqref{eq:scp_tau_sum_sup_bound} and using the union bound, we deduce that
\begin{align*}
    \frac1\tau \widehat\ct_{\tau, t}
    &
    \leq \frac{t - \tau}{t} \sup_{\theta \in \Theta} \left( \scp_\tau(\theta) - \frac12 \; \overline\scp_\tau(\theta) \right)
    + \frac1t \left( \scq_{\tau, t} - \overline\scq_{\tau, t} - \frac{3\varkappa}{4} \sum\limits_{s = \tau + 1}^t \left\|\Sigma^{1/2} \widehat\theta_{\tau, s - 1} \right\|^2 \right)
    \\&
    \leq \frac{3 e^B d}{\tau}
    + \frac{11 B \log(4 / \delta)}{4 \tau}
    + \frac{5 e^B \log(4 / \delta)}{2 \tau}
    + \frac{2}{t} \left(B + \frac{4 e^B}3 \right) \log(4 / \delta)
    \\&
    \leq \frac{3 e^B d}{\tau}
    + \frac{19 B \log(4 / \delta)}{4 \tau}
    + \frac{31 e^B \log(4 / \delta)}{6 \tau}
\end{align*}
with probability at least $(1 - \delta)$.
The proof is finished.

\myendproof

\subsection{Proof of Lemma \ref{lem:scp_uniform_bound}}
\label{sec:scp_uniform_bound_proof}

Throughout the proof, $X$ and $X'$ are i.i.d. random elements drawn from $\sfp$, which are independent of $X_1, \dots, X_\tau$.
Our strategy is to apply high-probability bounds based on local Rademacher complexity (see \citep[Theorem 3.3]{bbm05}) to the empirical process
\[
    \scp_\tau(\theta) = \frac{1}{\tau}\sum_{s = 1}^\tau f_\theta(X_s),
    \quad
    \theta \in \Theta,
\]
where
\[
    f_\theta(x) = -\log\left(\frac{1 + e^{-\theta^\top \psi(x)}}{2}\right) - \frac{1}{2}\theta^\top \psi(x)
    \quad \text{for any $\theta \in \Theta$.}
\]
First, let us check that
\begin{equation}
    \label{eq:scp_bernstein_condition}
    \Var\big( f_\theta(X) \big)
    \leq \frac{e^{2B} \varkappa^2}4 \left\|\Sigma^{1/2}\theta \right\|^2
    \leq -\frac{e^{2B} \varkappa}2 \; \overline\scp_\tau(\theta)
    \quad \text{for all $\theta \in \Theta$.}
\end{equation}
Indeed, let us introduce a function $h: \R \rightarrow \R$,
\[
    h(u) = -\log(1 + e^{-u}) - u/2
    \quad \text{for any $u \in \R$},
\]
and note that
\[
    |h'(u)|
    = \left| \frac{e^{-u}}{1 + e^{-u}} - \frac12 \right|
    = \frac12 \big| \tanh(-u/2)  \big|
    \leq \frac{e^{B} - 1}{2 (e^B + 1)}
    = \frac{e^{2B} - 1}{2 (e^B + 1)^2}
    \leq \frac{e^B \varkappa}2
\]
for any $u \in [-B, B]$. The constant $\varkappa$ is given by \eqref{eq:kappa_Sigma_def}. This yields that $h(u)$ is $(e^B \varkappa /2)$-Lipschitz on $[-B, B]$, and then
\begin{align*}
    \Var\big( f_\theta(X) \big)
    &
    = \frac12 \, \E \big( f_\theta(X) - f_\theta(X') \big)^2
    \\&
    \leq \frac{e^{2B} \varkappa^2}8 \, \E \big(\theta^\top\psi(X) - \theta^\top\psi(X') \big)^2
    \\&
    \leq \frac{e^{2B} \varkappa^2}4 \, \E \big(\theta^\top\psi(X) \big)^2
    = \frac{e^{2B} \varkappa^2}4 \left\|\Sigma^{1/2}\theta \right\|^2 .
\end{align*}
On the other hand, $h(u)$ is $\varkappa$-strongly concave on $[-B, B]$.
Then it holds that
\begin{align*}
    -\overline\scp_\tau(\theta)
    \geq -\overline\scp_\tau(0) - \nabla_\theta \overline\scp_\tau(0)^\top \theta + \frac{\varkappa}{2} 
    \, \E \big(\theta^\top\psi(X) \big)^2
    = \frac{\varkappa}{2} \, \E(\theta^\top\psi(X))^2
    = \frac{\varkappa}{2} \left\|\Sigma^{1/2}\theta \right\|^2,
\end{align*}
and \eqref{eq:scp_bernstein_condition} follows.

The inequality \eqref{eq:scp_bernstein_condition} means that we can apply Theorem 3.3 from \citep{bbm05} with the functional $T(f_\theta) = e^{2B} \varkappa^2 \|\Sigma^{1/2} \theta\|^2 / 4$. It only remains to bound the Rademacher complexity of the local star hull of $\{f_\theta : \theta \in \Theta\}$. For any $r > 0$, let
\[
    \Theta(r) = \left\{\theta \in \Theta : e^{2B} \varkappa^2 \, \|\Sigma^{1/2} \theta\|^2 / 4 \leq r \right\},
    \quad
    \cf(r) = \big\{f_\theta : \theta \in \Theta(r) \big\},
\]
and
\[
    \mathrm{star}\big( \cf(r) \big) = \big\{ \lambda f : f \in \cf(r), \, \lambda \in [0, 1] \big\}.
\]
Consider
\[
    \rr_\tau \left(\mathrm{star}\big( \cf(r) \big)\right)
    = \E \, \E_\sigma \sup\limits_{f \in \mathrm{star}( \cf(r))} \frac{1}{\tau} \sum_{s = 1}^\tau \sigma_s f(X_s),
\]
where $\sigma_1, \dots, \sigma_\tau$ are i.i.d. Rademacher random variables, which are independent of $X_1, \dots, X_\tau$.
Let us note that
\begin{align*}
    \rr_\tau \left(\mathrm{star}\big( \cf(r) \big)\right)
    &
    = \E \, \E_\sigma \sup\limits_{\lambda \in [0, 1], f \in \cf(r)} \frac{\lambda}{\tau} \sum_{s = 1}^\tau \sigma_s f(X_s)
    \\&
    \leq \E \, \E_\sigma \sup\limits_{f \in \cf(r)} \left| \frac{1}{\tau} \sum_{s = 1}^\tau \sigma_s f(X_s) \right|
    \\&
    = \E \, \E_\sigma \sup\limits_{\theta \in \Theta(r)} \left| \frac{1}{\tau} \sum_{s = 1}^\tau \sigma_s \left( \log\left(\frac{1 + e^{-\theta^\top \psi(X_s)}}{2}\right) + \frac{1}{2}\theta^\top \psi(X_s) \right) \right|.
\end{align*}
Since the function $h(u) = -\log(1 + e^{-u}) + \log 2 - u/2$ is $(e^B \varkappa/2)$-Lipschitz on $[-B, B]$ and $h(0) = 0$, we can apply the Talagrand contraction principle (see, for instance, \cite[Theorem 4.12]{lt91}) claiming that
\[
    \rr_\tau \left(\mathrm{star}\big( \cf(r) \big)\right)
    \leq \, \E \, \E_\sigma \sup\limits_{\theta \in \Theta(r)} \left| \frac{e^B \varkappa}{2\tau} \sum_{s = 1}^\tau \sigma_s \, \theta^\top \psi(X_s) \right|.
\]
The Cauchy-Schwartz inequality implies that
\begin{align*}
    \rr_\tau \left(\mathrm{star}\big( \cf(r) \big)\right)
    &
    \leq \E \, \E_\sigma \sup\limits_{\theta \in \Theta(r)} \frac{e^B \varkappa \|\Sigma^{1/2} \theta\|}{2\tau} \left\|\sum_{s = 1}^\tau \sigma_s \, \Sigma^{-1/2} \psi(X_s) \right\|
    \\&
    = \frac{\sqrt{r}}\tau \E \, \E_\sigma \left\|\sum_{s = 1}^\tau \sigma_s \, \Sigma^{-1/2} \psi(X_s) \right\|
    \\&
    \leq \frac{\sqrt{r}}\tau \left( \E \, \E_\sigma \left\|\sum_{s = 1}^\tau \sigma_s \, \Sigma^{-1/2} \psi(X_s) \right\|^2 \right)^{1/2}.
\end{align*}
The expectation in the right-hand side can be computed explicitly:
\begin{align*}
    \E \, \E_\sigma \left\|\sum_{s = 1}^\tau \sigma_s \, \Sigma^{-1/2} \psi(X_s) \right\|^2
    &
    = \E \, \E_\sigma \sum_{s = 1}^\tau \sum_{s' = 1}^\tau \sigma_s \sigma_{s'}  \psi(X_s)^\top \Sigma^{-1/2} \psi(X_{s'})
    \\&
    = \E \sum_{s = 1}^\tau \psi(X_s)^\top \Sigma^{-1} \psi(X_s)
    \\&
    = \tau d.
\end{align*}
Taking this identity into account, we finally obtain that
\[
    \rr_\tau \left(\mathrm{star}\big( \cf(r) \big)\right)
    \leq \sqrt{\frac{rd}\tau}
    \quad \text{for any $r > 0$.}
\]
According to \citep[Theorem 3.3]{bbm05}, for any $\delta \in (0, 1)$, with probability at least $(1 - \delta)$ simultaneously for all $\theta \in \Theta$ it holds that
\begin{equation}
    \label{eq:bbm_bound}
    -\overline{\scp}_\tau(\theta)
    \leq -2 \scp_\tau(\theta) + \frac{24 r^*}{e^{2B} \varkappa}
    + \frac{11 B \log(1 / \delta)}{2 \tau}
    + \frac{5 e^{2B} \varkappa \log(1 / \delta)}{\tau},
\end{equation}
where $r^*$ is the solution of the equation
\[
    r = \frac{e^{2B} \varkappa}{2} \sqrt{\frac{rd}\tau}.
\]
Substituting $r^* = e^{4B} \varkappa^2 d / (4 \tau)$ into \eqref{eq:bbm_bound}, we conclude that
\[
    -\overline{\scp}_\tau(\theta)
    \leq -2 \scp_\tau(\theta) + \frac{6 e^{2B} \varkappa d}{\tau}
    + \frac{11 B \log(1 / \delta)}{2 \tau}
    + \frac{5 e^{2B} \varkappa \log(1 / \delta)}{\tau}
\]
with probability at least $(1 - \delta)$ simultaneously for all $\theta \in \Theta$. The inequality $e^B \varkappa = e^{2B} / (1 + e^B)^2 \leq 1$ finishes the proof of the lemma.

\myendproof

\subsection{Proof of Lemma \ref{lem:scq_bound}}
\label{sec:scq_bound_proof}

The proof relies on the Bernstein inequality for martingales \citep{freedman75}. Throughout the proof, $X$ and $X'$ are i.i.d. random elements drawn from $\sfp$ independently of $X_1, \dots, X_t$.
Let us introduce
\[
    h(u) = -\log \frac{1 + e^u}2 + \frac{u}2
\]
and consider the martingale difference (with respect to the natural filtration)
\[
    \scv_s =
    \begin{cases}
        h\big( \psi(X_s)^\top \widehat{\theta}_{\tau, s - 1}) - \E_{X \sim \sfp} \, h\big( \psi(X_s)^\top \widehat{\theta}_{\tau, s - 1}),
        \quad \text{if $s \geq \tau + 1$;}\\
        0,
        \quad \text{otherwise.}
    \end{cases}
\]
Indeed, for any $s \in \mathbb N$, $\scv_s$ depends only on $X_1, \dots, X_s$ and $\E[\scv_{s + 1} \, | \, X_1, \dots, X_s] = 0$, because $\widehat\theta_{\tau, s}$ is a function of $X_1, \dots, X_s$.
We also note that $\scq_{\tau, t} - \overline{\scq}_{\tau, t} = \scv_{\tau + 1} + \ldots + \scv_t$.
Furthermore, since $h(u)$ is concave, $h(0) = 0$, and
\[
    |h'(u)|
    = \left|\frac12 - \frac{e^u}{1 + e^u} \right|
    = \left|\frac{e^u - 1}{2(e^u + 1)} \right|
    = \frac12 |\tanh(u/2)|
    \leq \frac{e^B - 1}{2 (e^B + 1)}
    = \frac{e^{2B} - 1}{2 (e^B + 1)^2}
    \leq \frac{e^B \varkappa}2
\]
for any $u \in [-B, B]$, where the constant $\varkappa$ is given by \eqref{eq:kappa_Sigma_def},
$\scv_s$ takes its values in $[-e^B \varkappa B / 2, 0] \subseteq [-B/2, 0]$ with probability $1$.
Let us elaborate on the conditional variance
\begin{align*}
    \Var[\scv_{s + 1} \, | \, X_1, \dots, X_s]
    = \frac{1}{2}\E_{X,X' \sim \sfp} \left( h\big( \psi(X)^\top \widehat{\theta}_{\tau, s}) -  h\big( \psi(X')^\top \widehat{\theta}_{\tau, s}) \right)^2.
\end{align*}
Using again the fact that $h(u)$ is $(e^B \varkappa / 2)$-Lipschitz on $[-B, B]$, we obtain that
\begin{align}
\label{eq:cond_var_bound}
    \notag
    \Var[\scv_{s + 1} \, | \, X_1, \dots, X_s]
    &\leq \frac{1}{2} \cdot \frac{e^{2B} \varkappa^2}{4} \E_{X, X' \sim \sfp} \left(\widehat{\theta}_{\tau, s}^\top \psi(X) - \widehat{\theta}_{\tau, s}^\top\psi(X') \right)^2
    \\&
    \leq \frac{e^B \varkappa}{4} \|\Sigma^{1/2}\widehat{\theta}_{\tau, s}\|^2.
\end{align}
In the last inequality we used the fact that $e^B \varkappa = e^{2B} / (1 + e^B)^2 \leq 1$ and that
\[
    \E_{X, X' \sim \sfp} \left(\widehat{\theta}_{\tau, s}^\top \psi(X) - \widehat{\theta}_{\tau, s}^\top\psi(X') \right)^2
    \leq 2 \, \E_{X \sim \sfp} \left(\widehat{\theta}_{\tau, s}^\top \psi(X) \right)^2.
\]
For each $t \geq \tau + 1$ and $a \leq b$ define the event
\begin{align*}
    \ce_t(a, b) = \left\{ a \leq \frac{12}{B e^B} \sum_{s = \tau + 1}^t \Var[\scv_s \, | \, X_1, \dots, X_{s - 1}] \leq b\right \} .
\end{align*}
On $\ce_t(a, b)$ we have that the sum of conditional variances of the scaled sequence $\{2 \scv_s / B : s \in \mathbb N\}$ satisfies
\begin{align}
\label{eq:cond_var_on_et}
    \frac{a e^B}{3 B}
    \leq \frac{4}{B^2}\sum_{s = \tau + 1}^t \Var[\scv_s \, | \, X_1, \dots, X_{s - 1}]
    \leq \frac{b e^B}{3 B} .
\end{align}
The peeling argument suggests that
\begin{align}
\label{eq:peeling_decomp}
    &\notag
    \p\left( \exists \, t \geq \tau + 1 \, : \, \frac{2( \scq_{\tau, t} - \overline{\scq}_{\tau, t})}{B} - \frac{3\varkappa}{2B}\sum_{s = \tau + 1}^t\|\Sigma^{1/2}\widehat{\theta}_{\tau, s - 1}\|^2  \geq \z \right)
    \\&
    \leq \p\left( \bigcup\limits_{t = \tau + 1}^\infty \left( \left\{ \frac{2( \scq_{\tau, t} - \overline{\scq}_{\tau, t})}{B} - \frac{3\varkappa}{2B}\sum_{s = \tau + 1}^t\|\Sigma^{1/2}\widehat{\theta}_{\tau, s - 1}\|^2  \geq \z \right\} \cap \ce_t(0, \z) \right) \right)
    \\&\notag
    + \sum_{k = 1}^\infty \p\left( \bigcup\limits_{t = \tau + 1}^\infty \left( \left\{ \frac{2( \scq_{\tau, t} - \overline{\scq}_{\tau, t})}{B} - \frac{3\varkappa}{2B}\sum_{s = \tau + 1}^t\|\Sigma^{1/2}\widehat{\theta}_{\tau, s - 1}\|^2  \geq \z \right\} \cap \ce_t(2^{k - 1} \z, 2^k \z) \right) \right).
\end{align}
Applying \cite[Theorem 4.1]{freedman75} and using \eqref{eq:cond_var_on_et}, we obtain that
\begin{align}
\label{eq:peeling_zero}
    &\notag
    \p\left( \bigcup\limits_{t = \tau + 1}^\infty \left( \left\{ \frac{2( \scq_{\tau, t} - \overline{\scq}_{\tau, t})}{B} - \frac{3\varkappa}{2B}\sum_{s = \tau + 1}^t\|\Sigma^{1/2}\widehat{\theta}_{\tau, s - 1}\|^2  \geq \z, \right\} \cap \ce_t(0, \z) \right) \right)
    \\&
    \leq \p\left( \bigcup\limits_{t = \tau + 1}^\infty \left( \left\{ \frac{2( \scq_{\tau, t} - \overline{\scq}_{\tau, t})}{B} \geq \z, \right\} \cap \ce_t(0, \z) \right) \right)
    \\&\notag
    \leq \exp\left\{ -\frac{\z^2}{2\z + 2 \z e^B / (3 B)}\right\}
    = \exp\left\{ -\frac{\z}{2 + 2 e^B / (3 B)}\right\}.
\end{align}
Similarly, for any $k \in \mathbb{N}$, we have that
\begin{align*}
    &
    \p\left( \bigcup\limits_{t = \tau + 1}^\infty \left( \left\{ \frac{2( \scq_{\tau, t} - \overline{\scq}_{\tau, t})}{B} - \frac{3\varkappa}{2B}\sum_{s = \tau + 1}^t\|\Sigma^{1/2}\widehat{\theta}_{\tau, s - 1}\|^2  \geq \z \right\} \cap \ce_t(2^{k - 1} \z, 2^k \z) \right) \right)
    \\&
    \leq \p\left( \bigcup\limits_{t = \tau + 1}^\infty \left( \left\{ \frac{2( \scq_{\tau, t} - \overline{\scq}_{\tau, t})}{B} \geq (1 + 2^{k - 2}) \z \right\} \cap \ce_t(2^{k - 1} \z, 2^k \z) \right) \right)
    \\&
    \leq \exp\left\{-\frac{(1 + 2^{k - 2})^2 \z^2}{2(1 + 2^{k - 2})\z + 2^{k + 1} \z e^B / (3B)}\right\}.
\end{align*}
Here we used the observation that, due to \eqref{eq:cond_var_bound},
\begin{align*}
    \frac{3\varkappa}{2B} \sum_{s = \tau + 1}^t \|\Sigma^{1/2} \widehat{\theta}_{\tau, s - 1} \|^2
    \geq \frac{6}{B e^B} \sum_{s = \tau + 1}^t \Var[\scv_s \, | \, X_1, \dots, X_{s - 1}]
    \geq 2^{k - 2} \z
    \quad \text{on $\ce_t(2^{k - 1}\z, 2^k\z)$.}
\end{align*}
Straightforward calculations imply that
\begin{align}
\label{eq:peeling_k}
    \notag
    &
    \p\left( \bigcup\limits_{t = \tau + 1}^\infty \left( \left\{ \frac{2( \scq_{\tau, t} - \overline{\scq}_{\tau, t})}{B} - \frac{3\varkappa}{2B}\sum_{s = \tau + 1}^t\|\Sigma^{1/2}\widehat{\theta}_{\tau, s - 1}\|^2  \geq \z \right\} \cap \ce_t(2^{k - 1} \z, 2^k \z) \right) \right)
    \\&\notag
    \leq \exp\left\{-\frac{(1 + 2^{k - 2})^2 \z^2}{2(1 + 2^{k - 2})\z + 2^{k + 1} \z e^B / (3B)}\right\}
    \\&
    \leq \exp\left\{-\frac{(1 + 2^{k - 2}) \z}{2 + 2^{k + 1} e^B / (3B) / (1 + 2^{k - 2})}\right\}
    \\&\notag
    \leq \exp\left\{-\frac{2^{k - 2} \z}{2 + 8 e^B / (3B)}\right\}
    = \exp\left\{-\frac{2^{k - 1} \z}{4 + 16 e^B / (3B)}\right\}
\end{align}
for any $\z > 0$. In view of \eqref{eq:peeling_decomp}, \eqref{eq:peeling_zero}, and \eqref{eq:peeling_k}, the choice $\z = (4 + 16 e^B / (3B) ) \log(3 / \delta)$ ensures that
\begin{align*}
    \p\left( \exists \, t \geq \tau + 1 \, : \, \frac{2( \scq_{\tau, t} - \overline{\scq}_{\tau, t})}{B} - \frac{3\varkappa}{2B}\sum_{s = \tau + 1}^t\|\Sigma^{1/2}\widehat{\theta}_{\tau, s - 1}\|^2  \geq \z \right)
    \leq \frac{\delta^2}{9} + \sum_{k = 1}^\infty(\delta / 3)^k
    \leq \delta .
\end{align*}
Therefore, we conclude that, with probability at least $(1 - \delta)$,
\[
    \scq_{\tau, t} - \overline{\scq}_{\tau, t} - \frac{3\varkappa}{4} \sum_{s = \tau + 1}^t \|\Sigma^{1/2}\widehat{\theta}_{\tau, s - 1}\|^2 \leq  \left(2B + \frac{8 e^B}3 \right) \log(3 / \delta)
\]
for all $t \geq \tau + 1$, thereby finishing the proof.

\myendproof

\section{Proof of Theorem \ref{th:rl_dd_falcon}}
\label{sec:th_rl_dd_falcon_proof}

The analysis of the running length of Algorithm \ref{th:rl_dd_falcon} is straightforward. Indeed, Theorem \ref{th:ftal_test_stat_upper_bound} and the union bound yield that 
\[
    \max\limits_{1 \leq t \leq T} \widehat\cs_t
    \leq \max\limits_{1 \leq t \leq T} \max\limits_{1 \leq \tau \leq t - 1} \widehat\ct_{\tau, t}
    \leq 3e^B d + \frac{19B}{4} \log \big(2T (T - 1) / \delta \big) + \frac{31 B}{6} \log \big(2T(T - 1) / \delta \big)
    = \z 
\]
with probability at least $(1 - \delta)$. In other words, Algorithm \ref{alg:fast_change_point} makes at least $T$ until the false alarm with high probability.

In the rest of the proof, we focus on to the detection delay bound. Let us introduce
\begin{align*}
    \theta^\circ \in \argmin_{\theta \in \Theta} \left\|\theta^\top \psi - \log(\sfp / \sfq) \right\|_{L_2((\sfp + \sfq) / 2)}.
\end{align*}
Since
\begin{align*}
    -\widehat\ct_{\tstar, t} + \max\limits_{\theta \in \Theta} \ct_{\tstar, t}(\theta)
    &
    = \frac{\tstar}{t}\sum_{s = 1}^t \left( \varphi_{\tstar, s}(\widehat\theta_{\tstar, s - 1}) - \min\limits_{\theta \in \Theta} \sum_{s = 1}^t \varphi_{\tstar, s}(\theta) \right)
    \\&
    =
    \begin{cases}
        \tstar \, \regret_\ca(t) / t,
        \quad \text{if $t > \tstar$,}\\
        0 \quad \text{otherwise,}
    \end{cases}
\end{align*}
it suffices to show that $\ct_{\tstar, t}(\theta^\circ)$ satisfies the inequality 
\[
    \ct_{\tstar, t}(\theta^\circ)
    \geq \z + \tstar \max\limits_{s \geq \tstar} \frac{\regret_\ca(s)}{s}
    = \z + \tstar \, \mar_{\tstar}
    \quad \text{whenever $t = \tstar + \tau_{\min}$}
\]
with high probability.
Now fix an arbitrary $\theta \in \Theta$ and provide an upper bound on the variance of the statistic $\ct_{\tstar, t}(\theta)$ given by \eqref{eq:t}.
Since $X_1, \dots, X_t$ are i.i.d. random elements, the variance of $\ct_{\tstar, t}(\theta)$ equals to
\begin{align*}
    &
    \frac{(t - \tstar)^2}{t^2} \sum\limits_{s = 1}^\tstar \Var\left[ \log\left( \frac{1 + e^{-\theta^\top \psi(X_s)}}2 \right) \right] + \frac{(\tstar)^2}{t^2} \sum\limits_{s = \tstar + 1}^t \Var\left[ \log\left( \frac{1 + e^{\theta^\top \psi(X_s)}}2 \right) \right]
    \\&
    = \frac{(t - \tstar)^2 \tstar}{t^2} \Var\left[ \log\left( \frac{1 + e^{-\theta^\top \psi(X_1)}}2 \right) \right] + \frac{(\tstar)^2 (t - \tstar)}{t^2} \Var\left[ \log\left( \frac{1 + e^{\theta^\top \psi(X_t)}}2 \right) \right].
\end{align*}
Hence, it holds that
\begin{align}
    \label{eq:var_ct}
    \Var\big( \ct_{\tstar, t}(\theta) \big)
    &\notag
    \leq \frac{(t - \tstar) \tstar}{t} \Var\left[ \log\left( \frac{1 + e^{-\theta^\top \psi(X_1)}}2 \right) \right]
    \\&\quad
    + \frac{\tstar (t - \tstar)}{t} \Var\left[ \log\left( \frac{1 + e^{\theta^\top \psi(X_t)}}2 \right) \right].
\end{align}
Let us elaborate on the first term in the right-hand side. Due to the Cauchy-Schwarz inequality, we have
\begin{align}
    \label{eq:var_cauchy-schwarz}
    \Var\left[ \log\left( \frac{1 + e^{-\theta^\top \psi(X_1)}}2 \right) \right]
    &\notag
    \leq 2 \left\| \log\left( \frac{1 + e^{-\theta^\top \psi}}2 \right) - \log \frac{\sfp + \sfq}{2\sfp} \right\|_{L_2(\sfp)}^2
    \\&\quad
    + 2 \E \log^2 \left(\frac{2 \sfp(X_1)}{\sfp(X_1) + \sfq(X_1)}\right).
\end{align}
Using the fact that $u \mapsto \log\big( (1 + e^{-u}) / 2 \big)$ is $1$-Lipschitz, we obtain that 
\begin{align}
    \label{eq:approx_bound}
    \left\| \log\left( \frac{1 + e^{-\theta^\top \psi}}2 \right) - \log \frac{\sfp + \sfq}{2\sfp} \right\|_{L_2(\sfp)}^2
    &\notag
    = \left\| \log\left( \frac{1 + e^{-\theta^\top \psi}}2 \right) - \log\left( \frac{1 + e^{-\log (\sfp / \sfq)}}2 \right) \right\|_{L_2(\sfp)}^2
    \\&
    \leq \left\| \theta^\top \psi - \log(\sfp / \sfq) \right\|_{L_2(\sfp)}^2.
\end{align}
To bound the last term in the right-hand side of \eqref{eq:var_cauchy-schwarz}, we invoke the following lemma.

\begin{Lem}
\label{lem:log_square_kl_bound}
    For any probability distributions on $\sfx$ with densities $\sfp$ and $\sfq$ it holds that
    \begin{align*}
        \E_{X \sim \sfp} \log^2\left(\frac{2\sfp(X)}{\sfp(X) + \sfq(X)}\right) \leq 2 (1 + \log 2) \KL\left(\sfp, \frac{\sfp + \sfq}{2}\right) .
    \end{align*}
    
\end{Lem}

The proof of Lemma \ref{lem:log_square_kl_bound} can be found in Appendix \ref{sec:lem_log_square_kl_bound_proof}.
Applying Lemma \ref{lem:log_square_kl_bound} and combining \eqref{eq:var_cauchy-schwarz} and \eqref{eq:approx_bound}, we deduce that
\begin{align*}
    \Var\left[ \log\left( \frac{1 + e^{-\theta^\top \psi(X_1)}}2 \right) \right]
    \leq 2 \left\| \theta^\top \psi - \log(\sfp / \sfq) \right\|_{L_2(\sfp)}^2
    + 4 (1 + \log 2) \KL\left(\sfp, \frac{\sfp + \sfq}{2}\right) .
\end{align*}
Using similar reasoning, one can also show that
\begin{align*}
    \Var\left[ \log\left( \frac{1 + e^{\theta^\top \psi(X_t)}}2 \right) \right]
    \leq 2 \left\| \theta^\top \psi - \log(\sfp / \sfq) \right\|_{L_2(\sfq)}^2
    + 4 (1 + \log 2) \KL\left(\sfq, \frac{\sfp + \sfq}{2}\right) .
\end{align*}
Therefore, substituting the derived bounds into \eqref{eq:var_ct} and using \eqref{eq:rho_theta}, we conclude that
\begin{align}
\label{eq:var_func_theta}
    \Var(\ct_{\tstar, t}(\theta))
    \leq \frac{(t - \tstar)\tstar}{t} \left( 4 \rho^2(\Theta) + 8 (1 + \log 2) \JS(\sfp, \sfq) \right) .
\end{align}
We are now ready to invoke the Bernstein inequality.
With probability at least $(1 - \delta)$, we have that
\begin{align*}
    \ct_{\tstar, t}(\theta^\circ) \geq \E \ct_{\tstar, t}(\theta^\circ) - \sqrt{2 \Var\big(\ct_{\tstar, t}(\theta^\circ) \big) \log(1 / \delta)} - 3B\log(1 / \delta) .
\end{align*}
Applying Lemma \ref{lem:js} and \eqref{eq:var_func_theta}, we obtain that
\begin{align*}
    \ct_{\tstar, t}(\theta^\circ)
    &
    \geq \frac{2 \tstar(t - \tstar)}{t} \left(\JS(\sfp, \sfq) - \frac{\rho^2(\Theta)}8 \right) - 3B \log(1 / \delta)
    \\&\quad
    - 2\sqrt{\frac{2\tstar (t - \tstar)}{t} \big(\rho^2(\Theta) + 4\JS(\sfp, \sfq) \big) \log(1 / \delta)}.
\end{align*}
By Young's inequality, we have that, with probability at least $(1 - \delta)$,
\begin{align*}
    &
    2\sqrt{\frac{2\tstar(t - \tstar)}{t} \big(\rho^2(\Theta) + 4\JS(\sfp, \sfq) \big) \log(1 / \delta)}
    \\&
    \leq \frac{2\tstar(t - \tstar)}{t}\left(\frac{\rho^2(\Theta)}{8} + \frac{\JS(\sfp, \sfq)}{2}\right) + 8\log(1 / \delta) .
\end{align*}
Thus, we conclude that
\begin{align*}
    \ct_{\tstar, t}(\theta^\circ)
    \geq \frac{\tstar(t - \tstar)}{t}\left(\JS(\sfp, \sfq) - \frac{\rho^2(\Theta)}{2}\right) - (3B + 8) \log(1 / \delta) ,
\end{align*}
with probability at least $(1 - \delta)$.
The condition $\tstar \geq \tau_{\min}$, together with the definition of $\tau_{\min}$, implies that for $t = \tstar + \tau_{\min}$ we have
\[
    \ct_{\tstar, t}(\theta^\circ)
    \geq \frac{\tau_{\min}}{2}\left(\JS(\sfp, \sfq) - \frac{\rho^2(\Theta)}{2}\right) - (3B + 8) \log(1 / \delta)
    \geq \zeta + \tstar \mar_\tstar,
\]
on an event of probability at least $(1 - \delta)$.
The proof is finished.

\myendproof

\subsection{Proof of Lemma \ref{lem:log_square_kl_bound}}
\label{sec:lem_log_square_kl_bound_proof}

Let us denote $\mathsf{r}(x) = (\sfp(x) + \sfq(x)) / 2$ for brevity.
Then, it follows that
\begin{align*}
    &\E_{X \sim \sfp} \log^2\left(\frac{2\sfp(X)}{\sfp(X) + \sfq(X)}\right)
    = \E_{X \sim \mathsf{r}}\left[ \frac{\sfp(X)}{\mathsf{r}(X)} \log^2\left(\frac{\sfp(X)}{\mathsf{r}(X)}\right) \right] .
\end{align*}
Taking into account that $\E_{X \sim \mathsf{r}}[\sfp(X) / \mathsf{r}(X)] = 1$, we also observe that
\begin{align*}
    \KL\left(\sfp, \frac{\sfp + \sfq}{2}\right) = \E_{X \sim \mathsf{r}}\left[\frac{\sfp(X)}{\mathsf{r}(X)}\left(\log\left(\frac{\sfp(X)}{\mathsf{r}(X)}\right) - 1\right) + 1 \right] .
\end{align*}
Since $0 \leq \sfp(x) / \mathsf{r}(x) \leq 2$, then it suffices to show that the function $f(x) = x \log x - x + 1 - c \, x \log^2 x$ is non-negative for all $0 \leq x \leq 2$, where $c^{-1} = 2(1 + \log 2)$.
Indeed, it holds that
\begin{align*}
    f'(x) = \log x - c (\log^2 x + 2 \log x),
    \quad f''(x) = x^{-1}\left( 1 - 2c - 2c \log x \right) .
\end{align*}
By the choice of $c$, we ensure that $1 - 2c - 2c \log x \geq 0$ for every $x \in [0, 2]$.
Therefore, the function $f$ is convex on $[0, 2]$ and achieves its minimal value $0$ at $x = 1$.
Consequently, $f$ is non-negative on $[0, 2]$, and the claim immediately follows.

\myendproof

\end{document}